\documentclass{article}

\usepackage[accepted]{want_icml2024}

\usepackage[accsupp]{axessibility}  

\usepackage[pagebackref,breaklinks,colorlinks]{hyperref}

\usepackage{amsmath,amsfonts,bm}

\def\eqref#1{equation~\ref{#1}}

\def\1{\bm{1}}
\newcommand{\train}{\mathcal{D}}

\def\vtheta{{\bm{\theta}}}
\def\vdelta{{\bm{\Delta}}}
\def\vepsilon{{\bm{\epsilon}}}

\def\vb{{\bm{b}}}
\def\vc{{\bm{c}}}

\def\vg{{\bm{g}}}

\def\vp{{\bm{p}}}

\def\vx{{\bm{x}}}

\def\mH{{\bm{H}}}

\DeclareMathAlphabet{\mathsfit}{\encodingdefault}{\sfdefault}{m}{sl}
\SetMathAlphabet{\mathsfit}{bold}{\encodingdefault}{\sfdefault}{bx}{n}

\usepackage{epsfig}
\usepackage{graphicx}
\usepackage{amsmath}
\usepackage{amssymb}

\usepackage{booktabs}

\usepackage[utf8]{inputenc} 
\usepackage[T1]{fontenc}    
\usepackage{url}            
\usepackage{amsfonts}       
\usepackage{nicefrac}       
\usepackage{microtype}      

\usepackage{breqn}
\usepackage{float}
\usepackage{caption}
\usepackage{subcaption}
\usepackage[cal=cm]{mathalfa}
\usepackage{comment}
\usepackage{mathtools}
\usepackage{bbm}

\usepackage{multirow}
\usepackage{threeparttable}
\usepackage{xcolor,colortbl}
\usepackage{pifont}
\usepackage{tabularray}
\usepackage{svg}

\newcommand{\ourcell}{\cellcolor[rgb]{1,0.808,0.576}}
\DeclarePairedDelimiter{\nint}\lfloor\rceil
\DeclareMathOperator{\tr}{tr}
\usepackage[capitalize,noabbrev]{cleveref}

\usepackage[textsize=tiny]{todonotes}

\icmltitlerunning{Fisher-aware Quantization for DETR Detectors with Critical-category Objectives}

\begin{document}

\twocolumn[
\icmltitle{Fisher-aware Quantization for DETR Detectors with\\ Critical-category Objectives}



\icmlsetsymbol{equal}{*}

\begin{icmlauthorlist}
\icmlauthor{Huanrui Yang}{equal,UCB}
\icmlauthor{Yafeng Huang}{equal,NJU}
\icmlauthor{Zhen Dong}{UCB}
\icmlauthor{Denis A Gudovskiy}{PAN}
\icmlauthor{Tomoyuki Okuno}{PAN}
\icmlauthor{Yohei Nakata}{PAN}
\icmlauthor{Yuan Du}{NJU}
\icmlauthor{Kurt Keutzer}{UCB}
\icmlauthor{Shanghang Zhang}{PKU}
\end{icmlauthorlist}

\icmlaffiliation{PKU}{School of Computer Science, Peking University}
\icmlaffiliation{UCB}{University of California, Berkeley}
\icmlaffiliation{NJU}{Nanjing University}
\icmlaffiliation{PAN}{Panasonic Holdings Corporation}

\icmlcorrespondingauthor{Huanrui Yang}{huanrui@berkeley.edu}
\icmlcorrespondingauthor{Shanghang Zhang}{shanghang@pku.edu.cn}

\icmlkeywords{DETR, quantization, critical-category, Fisher}

\vskip 0.3in
]



\printAffiliationsAndNotice{\icmlEqualContribution} 

\begin{abstract}
The impact of quantization on the overall performance of deep learning models is a well-studied problem. However, understanding and mitigating its effects on a more fine-grained level is still lacking, especially for harder tasks such as object detection with both classification and regression objectives. This work defines the performance for a subset of task-critical categories i.e. the critical-category performance, as a crucial yet largely overlooked fine-grained objective for detection tasks. We analyze the impact of quantization at the category-level granularity, and propose methods to improve performance for the critical categories. Specifically, we find that certain critical categories have a higher sensitivity to quantization, and are prone to overfitting after quantization-aware training (QAT). To explain this, we provide theoretical and empirical links between their performance gaps and the corresponding loss landscapes with the Fisher information framework. Using this evidence, we apply a Fisher-aware mixed-precision quantization scheme, and a Fisher-trace regularization for the QAT on the critical-category loss landscape. The proposed methods improve critical-category metrics of the quantized transformer-based DETR detectors. They are even more significant in case of larger models and higher number of classes where the overfitting becomes more severe. For example, our methods lead to 10.4\% and 14.5\% mAP gains for, correspondingly, 4-bit DETR-R50 and Deformable DETR on the most impacted critical classes in the COCO Panoptic dataset.
\end{abstract}

\section{Introduction}
\label{sec:intro}




Object detection is a challenging core application in computer vision, which is crucial for practical tasks such as autonomous driving. 
Recent DEtection TRansformer (DETR) model~\cite{carion2020end} and its variants achieve state-of-the-art results on multiple detection benchmarks~\cite{liu2022dab}. However, their performance comes at the cost of large model sizes and slow inference. Then, quantization~\cite{choi2018pact,dong2020hawq,dong2019hawq,polino2018model,yang2021bsq} is typically applied to reduce the memory footprint and inference latency time on cloud and edge devices~\cite{horowitz20141}. Inevitably, the perturbation of weights and activations introduced by the quantization process degrades the performance of floating-point models. Previous research on quantization~\cite{dong2019hawq,yang2021bsq,xiao2022csq} mainly focuses on a \textit{trade-off between the model size and the overall performance} (e.g., average accuracy for classification and mean average precision (mAP) for detection).
%

\begin{figure*}[t]
    \centering
    \includegraphics[width=\textwidth]{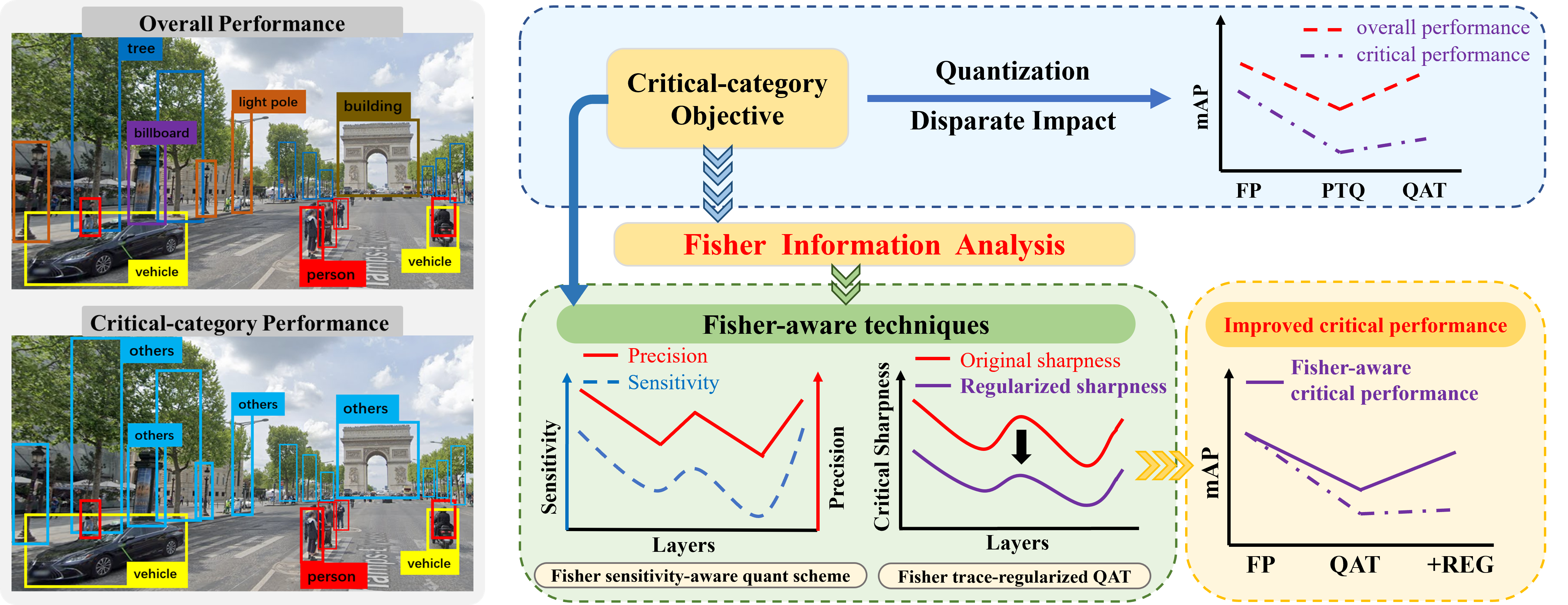}
    \caption{\textbf{Overview.} We investigate a practical setting with task-dependent critical-category objectives in~\cref{ssec:form}. We empirically observe disparate effects of quantization on the critical-category performance in~\cref{ssec:analy}, where post-training quantization (PTQ) and quantization-aware training (QAT) lead to performance gaps for critical categories w.r.t. to a floating-point (FP) model. We theoretically analyze such gaps using Fisher information framework and propose a Fisher-aware mixed-precision quantization scheme with regularization in~\cref{sec:method} to overcome these gaps for DETR models.}
    \label{fig:main}
\end{figure*}

However, a \textit{fine-grained performance objectives} are often more important than the overall performance in the real world~\cite{barocas-hardt-narayanan,tran2022pruning}. 
%
Suppose an autonomous vehicle is processing a scene containing people, vehicles, trees, light poles and buildings as illustrated in~\cref{fig:main}~(left)\footnote{Street scene photo in~\cref{fig:main} credits to Google Street View.}. Some non-critical objects (light poles, trees, and buildings) only need to be localized to avoid collision, yet misclassification within this group of categories is not as critical if they are all considered as ``other obstacles''. On the other hand, \textit{critical classes} such as a person or vehicle require both accurate classification and localization for a safe operation.
The overall performance cannot distinguish between an error within non-critical categories vs. a critical object error. In other words, it is missing granularity to represent the true task-critical objectives of real-world applications.
Yet to the best of our knowledge, for both post-training quantization (PTQ) and quantization-aware training (QAT), the analysis of the impact on such task-critical fine-grained objectives of object detection models is largely overlooked.

In this paper, we follow this practical yet neglected setting to formulate a set of task-critical objectives for DETR-based object detection models, accomplish a fine-grained quantization impact analysis, and propose techniques for improvements of the corresponding objectives. 
Specifically, we disentangle classification and localization objectives to define a fine-grained \textit{critical-category performance} with non-critical label transformation, as shown in the updated bounding boxes in~\cref{fig:main}~(left).
With this formulation, we provide a comprehensive analysis of the impact of quantization on the critical-category performance of DETR model. As illustrated in~\cref{fig:main}~(right), we find that quantization has a disparate effect on the category-wise performance, where some groups of classes are more sensitive to quantization with up to 1.7\% additional mAP drop for the public DETR object detector with ResNet-50 backbone. While QAT typically improves the overall performance, it can further increase performance gaps for the defined task-critical categories. We provide both theoretical and empirical analysis of such quantization effects using the loss surface landscape of the critical objectives by applying the Fisher information framework~\cite{old-fisher}.

Based on this analysis, we propose two novel techniques: Fisher-aware mixed-precision quantization scheme and Fisher-trace regularization. Both techniques optimize the landscape of critical objectives and, therefore, reduce overfitting and improve critical-category performance. Our experiments show consistent critical-category performance improvements for DETR object detectors with various backbones and architecture variants. The contributions of this paper are summarized as follows:
\begin{itemize}
    \item We formulate critical-category objectives for object detection and observe disparate effects of quantization on the performance of task-critical objectives.
    \item We provide analytical explanations of such quantization effects for DETR-based models using a theoretical link to the Fisher information matrix.
    \item Our Fisher-aware mixed-precision quantization scheme incorporates the sensitivity of critical-category objectives and increases their detection metrics. 
    \item Our Fisher-trace regularization further improves the loss landscape during quantization-aware training and the corresponding critical-category results. 
\end{itemize}

\section{Related Work}
\label{sec:background}

\noindent\textbf{Object detection.} Object detection is a core task for visual scene understanding. Conventional object detectors rely on a bounding box proposals~\cite{girshickICCV15fastrcnn}, fixed-grid anchors~\cite{redmon2016you} or window centers~\cite{tian2019fcos}. However, the performance of these methods is largely affected by bounding box priors and post-processing steps~\cite{carion2020end}. The transformer-based DETR~\cite{carion2020end} provides a fully end-to-end detection pipeline without surrogate tasks. Follow-up research further enhances DETR by introducing a deformable attention~\cite{zhu2020deformable}, query denoising~\cite{li2022dn}, and learnable dynamic anchors as queries~\cite{liu2022dab}. With the growing popularity of DETR-based architecture, we believe that understanding of quantization impact on DETR performance is an important topic, especially at the fine-grained level. Common object detection benchmarks evaluate fine-grained performance metrics that depend on the object size~\cite{lin2014microsoft} or occlusion status~\cite{Geiger2012CVPR}. However in practical applications, object type, i.e. its category, is often more important than the object size. This motivates us to further investigate detectors with critical-category objectives.

\noindent\textbf{Efficiency-performance tradeoff.} Multiple methods have been proposed to compress deep neural network (DNN) models, including pruning~\cite{han2015deep,wen2016learning,Yang2020DeepHoyer,yang2021nvit}, quantization~\cite{polino2018model,dong2020hawq,yang2021bsq,guo2022squant}, factorization~\cite{wen2017coordinating,ding2019centripetal,yang2020learning} and neural architecture search~\cite{wu2019fbnet,cai2019once}. In this work, we explore the impact of quantization that is widely supported by the hardware~\cite{horowitz20141} and can be almost universally applied to DNN model compression in architecture-agnostic fashion.

In previous work, average post-compression performance metrics are the key focus, but such overall performance hides the important fine-grained metrics e.g., the results for certain groups of categories. 
Recent works~\cite{tran2022pruning,good2022recall} analyze the disparate impact of pruning on classification accuracy, which leads to the fairness concerns~\cite{barocas-hardt-narayanan}. Our work extends this direction and investigates quantization effects of object detection at the critical-category performance granularity. 

\noindent\textbf{Second-order information in deep learning.} Unlike conventional optimization with the first-order gradients, recent research shows that the use of second-order information increases generalization and robustness of DNN models. Sharpness-aware minimization~\cite{foret2020sharpness} links a loss landscape sharpness with the model ability to generalize. The latter can be improved using a regularized loss with the Hessian eigenvalues~\cite{yang2021hero} computed w.r.t. parameter vector. Hessian eigenvalues are also used as importance estimates to guide the precision selection in mixed-precision quantization~\cite{dong2019hawq,dong2020hawq,yao2021hawq}. Given the difficulty of exact Hessian computation, Fisher information matrix is proposed as an approximation of the importance in pruning~\cite{kwon2022fast}. In this work, we link the quantization impact on the critical objectives with the second-order Fisher information. 

\section{Quantization Effects on Critical-Category Performance}
\label{sec:finegrain}
In this section, we introduce conventional training objectives for DETR in~\cref{ssec:pre}; formulate our critical-category objectives in~\cref{ssec:form}; and empirically analyze quantization effects on such critical-category performance in~\cref{ssec:analy}.

\subsection{Conventional DETR Training}
\label{ssec:pre}


Let $\vx$ be an input image from a dataset $\train$. Then, the DETR-type model $f_{\vtheta}(\cdot)$ with weights $\vtheta$ outputs a fixed-size set of $ N = K |\train|$ predictions $\hat{y}_i = \{ ( \hat{\vp}_i,\hat{\vb}_i ) \}_{i=1 \dots N}$, where $K$ is the model-dependent number of detections in each image, $\hat{\vp}_i$ is the vector of classification logits and $\hat{\vb}_i$ is the vector of bounding box coordinates. The former $\hat{\vp}_i \in \mathbb{R}^{C+1}$ contains logits for $C$ classes and an empty-box class ($\varnothing$). The predicted bounding box $\hat{\vb}_i \in \mathbb{R}^{4}$ consists of 4 scalars that define the center coordinates as well as the height and the width relative to the image size.

During the training, annotation is provided for each image in $\train$ as a set of ground truth objects $y_i = \{ (\vc_i, \vb_i) \}$, where $\vc_i$ is the one-hot vector with target class label and $\vb_i$ defines the bounding box. A Hungarian matching process is performed to find the closest one-to-one matching between ground truths and predictions including those with ``no object'' $\varnothing$ predictions. The training loss is computed between each pair of matched boxes, which is defined as a linear combination of a classification loss $\mathcal{L}_{cls}(\hat{\vp}_i, \vc_i)$ for all predictions, and a box loss $\mathcal{L}_{box}(\hat{\vb}_i, \vb_i)$ for all non-empty boxes.

The introduced notation is applicable to both the original DETR~\cite{carion2020end} and its more advanced variants such as DAB-DETR~\cite{liu2022dab}, Deformable DETR~\cite{zhu2020deformable} as well as any other detector with the end-to-end architecture.

\subsection{Proposed Critical-Category Objectives}
\label{ssec:form}

As discussed in~\cref{sec:intro}, the overall performance metric evaluated on the validation dataset is not the most effective objective in some real-world scenarios. Category-level fine-grained performance for some specific task-critical categories can be more crucial than the average metrics. Here we provide a practical definition of the critical-category objectives for the detection task, and a corresponding evaluation method when applied to DETR-type detectors.

In classification, class-level performance is often defined as the loss of the model on a subset of the validation dataset that contains objects from a certain group of classes~\cite{tran2022pruning}. However, such definition is not practical for object detection task, as each input image in the dataset contains multiple objects from different categories. Instead, this work defines the critical objective based on the entire validation dataset, but with a \textit{transformed model outputs} and annotations during the loss computation that focus detection towards a certain group of critical object categories.

Formally, assume there are in total $C$ categories in the dataset. Then, suppose the first $M$ categories are the critical ones for a certain task that requires both an accurate classification and localization. Hence, the rest of categories from $M+1$ to $C$ are non-critical and a misclassification between them is acceptable. This can be expressed by the transformed prediction $\tilde{\vp} \in \mathbb{R}^{M+2}$ as
\begin{equation}
\label{equ:logit_finegrain}
    \tilde{p}_j = \begin{cases}
    \hat{p}_j & j = 1 \dots M, \\
    \max \hat{p}_{M+1 \ldots C} & j = M+1, \\
    \hat{p}_{C+1} & j=M+2.
    \end{cases}
\end{equation}

The $(M+1)$-th category in $\tilde{\vp}$ corresponds to ``others'' which represents non-critical categories. The $\max$ function is used to avoid a distinction when classifying non-critical categories. The $(M+2)$-th category in $\tilde{\vp}$ is used for $\varnothing$ class which is originally defined as the $(C+1)$-th category in $\hat{\vp}$. 

The same~\cref{equ:logit_finegrain} transformation is also applied to the ground truth box label $\vc$, where all classes $c_{j \in \{M+1, \ldots, C\}}$ are redefined as the $(M+1)$-th label in the transformed $\tilde{\vc}$. No change is applied to the ground truth bounding boxes and the predicted boxes as we only define critical performance at the classification granularity to have a simplified yet practical and instructive setting.

The logit transformation can be applied directly to the output of a trained end-to-end model without any change to its architecture or weights $\vtheta$. The Hungarian matching, loss computation, and mAP computation can be performed without modification as well. We define the loss computed with the original $\hat{\vp}_i$ and $\vc_i$ as \textit{``overall objective''} and it is expressed as
\begin{equation}
\label{equ:loss_all}
    \mathcal{L}_A(\vtheta) = \frac{1}{N} \sum\nolimits_{i=1}^N \left( \mathcal{L}_{cls}(\hat{\vp}_i, \vc_i)+\mathcal{L}_{box}(\hat{\vb}_i, \vb_i) \right).
\end{equation}

Similarly, our \textit{``critical objective''} is defined with the transformed $\tilde{\vp}_i$ and $\tilde{\vc}_i$ as
\begin{equation}
\label{equ:loss_fine}
    \mathcal{L}_F(\vtheta) = \frac{1}{N} \sum\nolimits_{i=1}^N \left( \mathcal{L}_{cls}(\tilde{\vp}_i, \tilde{\vc}_i)+\mathcal{L}_{box}(\hat{\vb}_i, \vb_i) \right).
\end{equation}

Each objective in~\cref{equ:loss_all,equ:loss_fine} corresponds to either the \textit{``overall performance''} or the \textit{``critical performance''} when evaluating the mAP detection metric with the original or transformed outputs and labels, respectively.

\subsection{Empirical Evidence of Performance Gaps after Quantization}
\label{ssec:analy}

First, we empirically analyze how quantization affects the critical performance of a DETR model. We apply a symmetric linear quantizer $q(\cdot)$~\cite{dong2019hawq} to quantize weights $\vtheta$ to $Q$ bits of the pretrained DETR checkpoint with ResNet-50 backbone\footnote{\url{https://dl.fbaipublicfiles.com/detr/detr-r50-e632da11.pth}}, which can be expressed using the rounding operation $\nint{\cdot}$ as
\begin{equation}
\label{equ:quant}
    q(\vtheta) = \nint*{\frac{(2^{Q-1} - 1) \vtheta}{\max|\vtheta|}} \frac{\max|\vtheta|}{2^{Q-1} - 1}.
\end{equation}

\begin{table*}[tb]
\centering
\caption{Super-category mAPs before and after 4-bit uniform quantization, \%.}
\label{tab:fine_mAP_PTQ}
\resizebox{.8\linewidth}{!}{
\begingroup
    \setlength{\tabcolsep}{2pt}
\begin{tabular}{c|cccccccccccc|c} 
\toprule
Super category & Person & Vehicle & Outdoor & Animal & Accessory & Sports & Kitchen & Food  & Furniture & Electronic & Appliance & Indoor & Overall \\
\midrule
Pretrained  & 39.4  & 43.9   & 44.4   & 42.5  & 44.6     & 44.2  & 44.8   & 44.7 & 44.7     & 43.8     & 43.9     & 44.9  & 41.9   \\
PTQ 4-bit   & 20.1  & 23.3   & 23.9   & 22.3  & 23.9     & 23.7  & 24.2   & 23.9 & 23.7     & 23.4     & 23.5     & 24.0  & 20.9   \\
\midrule
mAP drop    & \textbf{19.3}  & 20.6   & 20.5   & 20.2  & 20.7     & 20.5  & 20.6   & 20.8 & \textbf{21.0}     & 20.4     & 20.4     & 20.9  & 21.0   \\
\bottomrule
\end{tabular}
\endgroup
}
\end{table*}
\begin{table*}[tb]
\centering
\caption{Super-category mAPs of 4-bit quantized model before and after QAT, \%.}
\label{tab:fine_mAP_QAT}
\resizebox{.8\linewidth}{!}{
\begingroup
    \setlength{\tabcolsep}{2pt}
\begin{tabular}{c|cccccccccccc|c} 
\toprule
Super category & Person & Vehicle & Outdoor & Animal & Accessory & Sports & Kitchen & Food  & Furniture & Electronic & Appliance & Indoor & Overall \\
\midrule
PTQ 4-bit   & 20.1  & 23.3   & 23.9   & 22.3  & 23.9     & 23.7  & 24.2   & 23.9 & 23.7     & 23.4     & 23.5     & 24.0  & 20.9   \\
QAT 4-bit   & 34.6  & 38.6   & 39.2   & 37.2  & 39.4     & 38.9  & 39.5   & 39.3 & 39.2     & 38.6     & 38.6     & 39.6  & 36.7   \\
\midrule
mAP gain    &  \textbf{14.5}  & 15.3   & 15.3   & 14.9  & 15.5     & 15.2  & 15.3   & 15.4 & 15.5     & 15.2     & 15.1     &  \textbf{15.6}  & 15.8   \\
\bottomrule
\end{tabular}
\endgroup
}
\end{table*}

We quantize all trainable weights in the DETR model using \cref{equ:quant} with an exception of the final feed-forward (FFN) layers for the class and bounding box outputs. Quantization of these FFN layers leads to a catastrophic performance drop in the PTQ setting~\cite{yuan2022ptq4vit}. A 4-bit quantization is applied uniformly to the weights of all layers for all experiments in this empirical study.

Without loss of generality, we define critical categories based on the ``super category'' labels in the COCO dataset~\cite{lin2014microsoft}. In total, 12 super categories are available in the COCO, where each contains from 1 to 10 categories of similar objects. 
For each selected super category, we consider all the categories within it as critical categories, while the rest of categories as non-critical and transform their logits and labels accordingly. The mAP measured at the transformed output is denoted as the critical mAP of this super category. 
For example, when measuring the critical performance of ``indoor'' super category, ``book'', ``clock'', ``vase'', ``scissors'', ``teddy bear'', ``hair drier'', and ``toothbrush'' are considered as critical categories (the first $M$ categories in the~\cref{equ:logit_finegrain} logit-label transformation), while others are set as non-critical.
We perform such evaluation for all 12 super categories to understand the category-level impact of DETR quantization.

As shown in~\cref{tab:fine_mAP_PTQ}, quantization has a disparate impact on the critical performance of the DETR model. The mAP drop after quantization has an up to 1.7\% gap. 
We further perform 50 epochs of QAT and report the critical performance in~\cref{tab:fine_mAP_QAT}. The performance increases differently for each super category with a gap of up to 1.1\% mAP. 

\section{Proposed Methods to Overcome Quantization Gaps}
\label{sec:method}

In this section, we theoretically analyze the causes of empirical performance gaps in~\cref{ssec:cause}. Then, we propose our methods to improve such performance from the aspect of quantization scheme design and quantization-aware training objective in~\cref{ssec:PTQ} and~\cref{ssec:QAT}, respectively. 

\subsection{Causes of Performance Gaps after Quantization}
\label{ssec:cause}

We investigate how quantization affects the critical objective $\mathcal{L}_F(\vtheta)$ for a pretrained DETR model with $\vtheta$ weights. We obtain the following theoretical results.

\noindent\textbf{Claim 1: Quantization-induced weight perturbation causes a larger Fisher trace of critical objectives and, therefore, inferior maximum likelihood estimates.}
The quantization process replaces the floating-point weights $\vtheta$ of the pretrained DETR model with the quantized values $q(\vtheta)$ using~\cref{equ:quant}. Effectively, this perturbs the weights away from their optimal values, which leads to an increase in the critical objective value. With the second-order Taylor expansion around $\vtheta$, the quantization-perturbed loss $\mathcal{L}_F(q(\vtheta))$ can be approximated using the non-perturbed objective $\mathcal{L}_F(\vtheta)$ as
\begin{equation}
\label{equ:Taylor}
    \mathcal{L}_F(q(\vtheta)) \approx \mathcal{L}_F(\vtheta) + \vg^T \vdelta + \vdelta^T \mH \vdelta / 2,
\end{equation}
where the gradient $\vg = \mathbb{E} \left[ \partial \mathcal{L}_F(\vtheta) / \partial \vtheta \right]$, the Hessian $\mH = \mathbb{E} \left[ \partial^2 \mathcal{L}_F(\vtheta) / ( \partial \vtheta \partial \vtheta^T) \right]$ and the weight perturbation or the quantization error $\vdelta = q(\vtheta)-\vtheta$.

Assuming the pretrained model converges to a local minimum, the first-order term can be ignored because $\vg \to 0$~\cite{lecun1989optimal} and~\cref{equ:Taylor} can be rewritten as
\begin{equation}
\label{equ:Taylor_sim}
    \mathcal{L}_F(\vtheta) - \mathcal{L}_F(q(\vtheta)) \propto - \vdelta^T \mH \vdelta.
\end{equation}

For large models such as DETR, computation of the exact Hessian matrix $\mH$ is practically infeasible. Previous research~\cite{kwon2022fast} shows that the Hessian estimate can be derived as the negative of Fisher information matrix $\mathcal{I}$ by
\begin{equation}
\label{equ:fisher}
    \mH = - \mathcal{I} = - \mathbb{E} \left[ \frac{\partial \mathcal{L}_F(\vtheta)}{\partial \vtheta}  \frac{\partial \mathcal{L}_F(\vtheta)}{\partial \vtheta^T} \right].
\end{equation}

Alternatively, we can interpret~\cref{equ:fisher} for a discriminative model $f_\vtheta(\vx)$ from~\cref{sec:finegrain} that maximizes $\log$-likelihood of the $p (y | \vx, \vtheta)$ density function using the empirical dataset $\train$ with the loss  $\mathcal{L}_F(\vtheta)$~\cite{Gudovskiy_2021_CVPR} as
\begin{equation}
\begin{split}
    \label{equ:fisher2}
    \mathcal{I} &= \mathbb{E}_{\train} \left[ \frac{\partial \mathcal{L}_F(\vtheta)}{\partial \vtheta}  \frac{\partial \mathcal{L}_F(\vtheta)}{\partial \vtheta^T} \right] \\ &= \frac{1}{N} \sum\nolimits_{i=1}^N  \left( \frac{\partial \log p (y_i | \vx_i)}{\partial \vtheta} \frac{\partial \log p (y_i | \vx_i)}{\partial \vtheta^T} \right).
\end{split}
\end{equation}

In practice, we can assume $\mathcal{I}$ to be diagonal~\cite{fim}, which simplifies~\cref{equ:Taylor_sim} to
\begin{equation}
\begin{split}
    \label{equ:sen}
    &\mathcal{L}_F(\vtheta) - \mathcal{L}_F(q(\vtheta)) \propto \vdelta^T \mathcal{I} \vdelta \\ &= \sum\nolimits_i \vdelta^2_i \left\Vert \partial \mathcal{L}_F(\vtheta) / \partial \vtheta_i \right\Vert_2^2 = \sum\nolimits_i \vdelta^2_i \mathcal{I}_{ii},
\end{split}
\end{equation}
where the latter result represents a sum of Fisher trace elements $\left( \tr(\mathcal{I}) = \sum\nolimits_i \mathcal{I}_{ii} \right)$ weighted by the squared quantization error over each $i$-th element of $\vtheta$.

\cref{equ:sen} provides a feasible yet effective sensitivity metric to estimate the impact of quantization noise. 
It analytically connects the quantization-induced weight perturbation with the maximum likelihood estimation in~\cref{equ:fisher2} for critical objectives using Fisher information framework~\cite{LY201740}.
Hence, an objective with larger sensitivity leads to inferior maximum likelihood estimates, i.e. the critical performance. 


\noindent\textbf{Claim 2: Sharp loss landscape leads to a poor test-time generalization for critical categories after quantization-aware training.} 
During the conventional QAT process, weights of the DETR model are trained to minimize the overall objective $\mathcal{L}_A(q(\vtheta))$. Nevertheless, a convergence of $\mathcal{L}_A$ does not guarantee good performance on all critical objectives $\mathcal{L}_F$. 
When compared to the overall objective, the critical objective with the focus on a subset of classes can be quickly minimized by the model during the training process which leads to a tendency of overfitting. 
The overfitting phenomenon is more severe with a larger model or with more classes in the overall training task.

To better analyze the issue of overfitting, we refer to the previous work on loss landscape sharpness~\cite{foret2020sharpness}, which finds a positive correlation between the generalization gap of the objective $\mathcal{L}_F$ and the sharpness $\mathcal{S}$ of the loss landscape around the local minima $q(\vtheta)$ of the QAT. The minima sharpness $\mathcal{S}(q(\vtheta))$ of the quantized model can be estimated as
\begin{equation}
\label{equ:sharp}
    \mathcal{S}(q(\vtheta)) = \max_{\|\vepsilon\|_2\leq\rho}\mathcal{L}_F(q(\vtheta) + \vepsilon) - \mathcal{L}_F(q(\vtheta)),
\end{equation}
where $\rho>0$ is a $\ell_2$ norm bound for the worst-case weight perturbation $\vepsilon$.

Finding the exact solution to the maximization in~\cref{equ:sharp} can be computationally costly. With the details in~\cref{ap:derive}, we can simplify it as  
\begin{equation}
\label{equ:grad_norm}
    \mathcal{S} \approx \max_{||\vepsilon||_2\leq\rho} \vepsilon^T \partial \mathcal{L}_F(q(\vtheta)) / \partial \vtheta
    \propto \tr(\mathcal{I}).
\end{equation}

Hence, the trace of the diagonal Fisher information matrix approximates the sharpness of the critical loss landscape for the quantized model. Sharp loss landscape leads to inferior test-time critical-category performance after QAT. 

\subsection{Fisher-aware Mixed-Precision Quantization Scheme}
\label{ssec:PTQ}

With the derived quantization impact on the loss in~\cref{equ:sen}, we propose a mixed-precision quantization scheme that minimizes the quantization effects within a model-size budget, which is defined as
\begin{equation}
\begin{split}
    \label{equ:ILP}
    \min_{Q_{1:L}} &\sum\nolimits_{i=1}^L \vdelta_i^2 \left\lVert \partial \left( \alpha \mathcal{L}_A(\vtheta) + \mathcal{L}_F(\vtheta) \right) / \partial \vtheta_i \right\rVert_2^2,\ \\ &\text{s.t.} \sum\nolimits_{i=1}^L Q_i \left\lVert \vtheta_i\right\rVert_0 \leq B,
\end{split}
\end{equation}
where $\vtheta$ is the weight vector of all $L$ layers in the model, $\vtheta_i$ is its $i$-th layer subset, and $\vdelta_i = q(\vtheta_i)-\vtheta_i$ is the $i$-th layer's quantization error when quantized to $Q_i$ bits. The budget $B$ is the model size allowance. The optimization problem in~\cref{equ:ILP} can be efficiently solved as an Integer Linear Programming (ILP) problem~\cite{dong2020hawq, yao2021hawq} with the discrete integer values for quantization precision $Q_i$.

Note that in~\cref{equ:ILP} we employ the Fisher information of both critical $\mathcal{L}_F$ and overall $\mathcal{L}_A$ objectives. This approach achieves good overall performance and increases the critical performance of interest. A hyperparameter $\alpha$ balances $\mathcal{L}_F$ and $\mathcal{L}_A$, which is selected using empirical cross-validation. 

\subsection{Fisher Trace Regularization for Quantization-aware Training}
\label{ssec:QAT}

Previous line of work on Sharpness-aware Minimization (SAM)~\cite{foret2020sharpness,liu2021sharpness} directly optimizes the sharpness estimate from~\cref{equ:sharp} by adding the worst-case weight perturbation in the training. However, we find that the complicated DETR architecture and its objective lead to a poor convergence for SAM-based methods. Moreover, a case with several critical objectives would involve multiple rounds of weight perturbation. Hence, this approach with explicit weight perturbation leads to optimization that is not scalable in our setup. 


Instead, to minimize loss sharpness $\mathcal{S}(q(\vtheta))$ during the DETR QAT optimization, we propose to follow the implicit sharpness derivation in~\cref{equ:grad_norm}. 
Specifically, for a critical objective $\mathcal{L}_F$, we add the Fisher trace regularization as
\begin{equation}
\label{equ:obj}
    \min_\vtheta \mathcal{L}_A(q(\vtheta))+\lambda \tr(\mathcal{I}_F),
\end{equation}
where $\lambda\geq 0$ is the strength of the regularization, and $\mathcal{I}_F$ denotes the Fisher information matrix of the critical objective $\mathcal{L}_F(q(\vtheta))$ w.r.t. weights $\vtheta$.

In addition to the DETR training loss terms in~\cref{equ:loss_all,equ:loss_fine}, we further add a distillation loss~\cite{hinton2015distilling} between the quantized (student) model and the pretrained full-precision (teacher) model to follow a common QAT practice~\cite{dong2020hawq,yang2021bsq}. The distillation objective consists of a KL-divergence loss for class logits of the student and teacher models, and a $\ell_1$ loss for the corresponding bounding box coordinates. Since we expect the student model to have the same behavior as the teacher model, the distillation loss uses a fixed one-to-one mapping between the predicted boxes of the two models without performing the Hungarian matching.
\section{Experiments}
\label{sec:exp}

\subsection{Experimental Setup}
\label{sec:exp_setup}

\noindent\textbf{Datasets and metrics.} We follow DETR~\cite{carion2020end} setup and use two variants of the COCO 2017 dataset~\cite{lin2014microsoft}: COCO detection and COCO panoptic segmentation. The detection dataset contains 118K training images and labels with 80 categories combined into 12 super categories. The panoptic dataset consists of 133K training examples and corresponding labels with 133 categories and 27 super categories. Both variants contain 5K data points in the validation set, which we use to evaluate both the overall and critical mAP in our experiments. Additional CityScapes~\cite{cityscapes} dataset evaluations are reported in~\cref{sec:PTQ_ablation}.

We follow~\cref{ssec:analy} and define the critical mean average precision (mAP) for each super category by considering all classes within it as critical while the rest of classes are non-critical. All mAPs reported in the tables are in percentage points. In case of COCO panoptic dataset, we report the box detection mAP$_{\textrm{box}}$.

\noindent\textbf{Model architectures.} We conduct the majority of our experiments on the DETR model with ResNet-50 backbone (DETR-R50). To show the scalability, we also experiment with larger ResNet-101 backbone (DETR-R101) and more advanced architectures such as DAB-DETR~\cite{liu2022dab} and Deformable DETR~\cite{zhu2020deformable}.

\noindent\textbf{Implementation details.} 
We perform quantization of the pretrained models using their public checkpoints. We apply symmetric layer-wise weight quantization using~\cref{equ:quant}, where weights are scaled by the $\max |\vtheta|$ without clamping. We keep normalization and softmax operations at full precision. We compute Fisher trace for our method using all training set for sensitivity analysis. But for implementation of HAWQ-V2~\cite{dong2020hawq} baseline, we randomly sample 1,000 training images due to high computational cost of Hessian estimation. We solve mixed-precision quantization problem in~\cref{equ:ILP} by the ILP with 3- to 8-bit budget $B$ for each layer. 
We perform QAT with the straight-through gradient estimator~\cite{bengio2013estimatingSTE} for 50 epochs with 1e-5 learning rate. 
Regularization strength $\lambda$ in~\cref{equ:obj} grows linearly from 1e-3 to 5e-3 throughout the training when our Fisher regularization is applied. In all experiments we report the mean and, if shown, $\pm$ standard error of the final 5 epochs of training to mitigate training variance.

\subsection{Quantitative Results of Fisher-aware Quantization}



\begin{table*}[t]
\centering
\caption{Critical-category mAP after PTQ on COCO detection dataset, \%. Our Fisher-critical scheme with the fine-grained objective surpasses others.}
\label{tab:PTQ_COCO_fine}
\small
\begin{tabular}{cc|ccc|ccc}
\toprule
\multirow{2}{*}{Model} & \multirow{2}{*}{\shortstack{Quant. \\ scheme}} & \multicolumn{3}{c|}{4-bit} & \multicolumn{3}{c}{6-bit} \\
&  & Person~  &  Animal~ & Indoor &  Person~   &  Animal~ & Indoor \\
\midrule
\multirow{3}{*}{DETR-R50} 
& Uniform & 34.6                       & 37.2             & 39.6            & 37.3                        & 40.0             & 42.4           \\       
& HAWQ-V2    & 35.31\tiny$\pm$0.1           & 37.90\tiny$\pm$0.2            & 40.20\tiny$\pm$0.2                       & 37.29\tiny$\pm$0.0             & 40.20\tiny$\pm$0.1           & 42.60\tiny$\pm$0.1 \\ 
& Fisher-overall                     & 35.35\tiny$\pm$0.0                         & 37.96\tiny$\pm$0.2            & 40.20\tiny$\pm$0.2            & 37.58\tiny$\pm$0.1                          & 40.74\tiny$\pm$0.1            & 43.10\tiny$\pm$0.1            \\
&\ourcell Fisher-critical                 &\ourcell \textbf{35.56}\tiny$\pm$0.1                          &\ourcell \textbf{38.10}\tiny$\pm$0.1             &\ourcell \textbf{40.33}\tiny$\pm$0.0           &\ourcell \textbf{37.73}\tiny$\pm$0.0                         &\ourcell \textbf{40.86}\tiny$\pm$0.1          &\ourcell \textbf{43.26}\tiny$\pm$0.1            \\
\midrule
\multirow{2}{*}{DETR-R101}
& Fisher-overall                     & 36.36                          & \textbf{39.30}             & 41.70            & 39.1                          & 42.0             & 44.4            \\
&\ourcell Fisher-critical                 & \ourcell\textbf{36.42}                          & \ourcell 39.23             &\ourcell \textbf{41.80}            & \ourcell\textbf{39.2}                          &\ourcell \textbf{42.5}             &\ourcell \textbf{44.9}            \\
\midrule
\multirow{3}{*}{\shortstack{DAB\\DETR-R50}}
& Uniform & 22.32                         & 25.68          & 27.60          & 26.24                        & 29.76            & 31.88          \\
& HAWQ-V2 & 8.26            & 11.66          & 12.80         & 19.10                  & 19.90            & 21.60            \\
& Fisher-overall                     & 22.82                         & 27.02             & \textbf{28.96}          & 26.06                      & 29.20          & 31.32          \\
&\ourcell Fisher-critical                 & \ourcell \textbf{23.18}                        &\ourcell \textbf{27.86}            & \ourcell 27.98            &\ourcell \textbf{26.38}                         &\ourcell \textbf{29.28}           & \ourcell \textbf{31.88}         \\
\midrule
\multirow{3}{*}{\shortstack{Deformable\\DETR-R50}}
& Uniform & 28.9                          & 32.8             & 34.3            & 46.0                          & 49.1             & 51.4            \\
& Fisher-overall                     & 42.7                          & 46.2             & 48.4            & 46.3                          & \textbf{49.5}             & 51.8            \\
&\ourcell Fisher-critical                 &\ourcell \textbf{43.1}                          &\ourcell \textbf{46.3}             &\ourcell \textbf{48.8}            &\ourcell \textbf{46.6}                          &\ourcell \textbf{49.5}             &\ourcell \textbf{52.0}            \\
\bottomrule   
\end{tabular}
\vspace{-10pt}
\end{table*}

\begin{table*}[t]
\centering
\caption{Critical-category mAP$_{\textrm{box}}$ for various post-training quantization (PTQ) schemes on COCO panoptic dataset, \%. Our Fisher-critical scheme exceeds others.}
\label{tab:PTQ_pano_fine}
\small
\begin{tabular}{cc|ccc|ccc}
\toprule
\multirow{2}{*}{Model} & \multirow{2}{*}{\shortstack{Quant. \\ scheme}} & \multicolumn{3}{c|}{4-bit} & \multicolumn{3}{c}{5-bit} \\
&  & Person~  &  Animal~ & Indoor &  Person~   &  Animal~ & Indoor \\
\midrule

\multirow{3}{*}{DETR-R50}
& Uniform                           & 8.5                           & 11.4             & 12.4            & 8.9                          & 13.7             & 16.0            \\
& Fisher-overall                     & 16.64                           & 21.60             & 23.80             & 18.79                         & 24.00             & 26.70            \\
&\ourcell Fisher-critical                         &\ourcell \textbf{16.68}             &\ourcell \textbf{21.69}            & \ourcell\textbf{23.85}         & \ourcell \textbf{19.05 }               &\ourcell \textbf{24.15 }           & \ourcell\textbf{26.87}            \\
\bottomrule   
\end{tabular}
\vspace{-10pt}
\end{table*}

We compare the proposed Fisher-aware mixed-precision quantization scheme from~\cref{ssec:PTQ} with the linear uniform quantization~\cite{polino2018model} and the mixed-precision HAWQ-V2~\cite{dong2020hawq} baselines. 
\cref{tab:PTQ_COCO_fine,tab:PTQ_pano_fine} report the critical mAP of super category ``person'', ``animal'', and ``indoor'' for the COCO detection and panoptic segmentation datasets, respectively.
We apply the baselines and our Fisher-overall variant for ablation study with only the overall objective $\mathcal{L}_A$, and evaluate quantized models on the selected super categories. Similarly, we report results for the proposed Fisher-critical scheme with the fine-grained $\mathcal{L}_F$ objective.

With the same mixed-precision quantization budget, our Fisher-aware method consistently outperforms uniform quantization and the mixed-precision scheme derived from HAWQ-V2 on different models and datasets. We note that the improvement of the HAWQ-V2 over the uniform quantization is not consistent on DETR-based models. This is caused by the instability of Hessian trace estimation for the complex DETR architecture and the harder object detection task. Fisher-aware approach, on the other hand, is stable. In addition, we compare the time to estimate the Fisher and the Hessian traces for a batch of images on P100 GPU and find that the Fisher trace can be estimated with $200-300\times$ \textit{less latency} than the Hessian one. This allows us to estimate Fisher trace with a large amount of training data, which leads to a higher accuracy and stability.

Quantitatively, our Fisher-critical scheme on COCO detection dataset improves critical mAP by up to 0.2\% for DETR-R50, 0.5\% for DETR-R101, 0.8\% for DAB DETR-R50, and 0.4\% for Deformable DETR-R50, respectively.
With more categories in the panoptic dataset, the impact of quantization on each individual category becomes even higher. Fisher-aware quantization variant with the overall objective only ($\mathcal{L}_A$) improves critical mAP by about 2$\times$ over the uniform quantization baseline. Further improvement on critical mAP is consistently achieved with the proposed Fisher-critical quantization scheme that incorporates the fine-grained objective $\mathcal{L}_F$. These results shows the \textit{importance of the proposed scheme with critical objectives} when applying object detection models to real-world applications. Moreover, the common overall mAP metric is not significantly affected when using the conventional $\mathcal{L}_A$ objective or the proposed scheme with $\mathcal{L}_F$ as additionally evaluated in the~\cref{sec:PTQ_ablation}.

\subsection{Quantitative Results for QAT with Fisher-trace Regularization}

\begin{table}[t]
\centering
\caption{Quantization-aware training (QAT) results of the overall performance and for the "person" critical category using the DETR-R50 model on COCO detection (mAP) and panoptic (mAP$_{\textrm{box}}$) datasets, \%. Fisher-critical quantization is applied before QAT.}
\label{tab:QAT_coco}
\resizebox{\linewidth}{!}{
\begingroup
    \setlength{\tabcolsep}{1pt}
\begin{tabular}{cc|cc|cc}
\toprule
\multirow{2}{*}{Model} & \multirow{2}{*}{\shortstack{QAT \\ objective}} & \multicolumn{2}{c|}{4-bit} & \multicolumn{2}{c}{6-bit det. / 5-bit panoptic} \\
&  & Overall~  &  Person~ & Overall~ &  Person~  \\
\midrule
\multirow{2}{*}{\shortstack{DETR-R50 \\ (detection)}} 
& Overall                     & \textbf{37.07}\tiny$\pm$0.07 & 35.56\tiny$\pm$0.08  & 39.67\tiny$\pm$0.10 & 37.73\tiny$\pm$0.02  \\
&\ourcell Fisher reg.         &\ourcell 36.97\tiny$\pm$0.06 & \ourcell\textbf{35.75}\tiny$\pm$0.04   &\ourcell \textbf{39.70}\tiny$\pm$0.08 & \ourcell\textbf{37.78}\tiny$\pm$0.01                      \\

\midrule
\multirow{2}{*}{\shortstack{DETR-R50 \\ (panoptic)}} 
& Overall                     & 33.24\tiny$\pm$0.10 & 16.68\tiny$\pm$0.01 & 36.08\tiny$\pm$0.07 & 19.05\tiny$\pm$0.09   \\
&\ourcell Fisher reg.              &\ourcell \textbf{33.29}\tiny$\pm$0.05 & \ourcell\textbf{16.79}\tiny$\pm$0.03 &\ourcell \textbf{36.12}\tiny$\pm$0.06 &\ourcell \textbf{19.39}\tiny$\pm$0.11   \\

\bottomrule
\end{tabular}
\endgroup
}
\vspace{-10pt}
\end{table}


\cref{tab:QAT_coco} compares the post-QAT results when using the conventional overall loss $\mathcal{L}_A$ only vs. our approach with Fisher-trace regularization from~\cref{ssec:QAT} on the COCO detection and panoptic datasets. 
The experimental results show that the proposed regularization further improves critical-category metrics.

When combined with the mixed-precision quantization scheme from~\cref{ssec:PTQ}, our method on COCO detection dataset (\cref{tab:QAT_coco}~(top)) leads to a 1.15\% and 0.48\% critical ("person" class) performance improvement for DETR-R50 model over the uniform quantization in~\cref{tab:PTQ_COCO_fine} for 4-bit (34.6\% $\rightarrow$ 35.75\%) and 6-bit (37.3\% $\rightarrow$ 37.78\%) precision, respectively. Note that our regularization scheme has a negligible impact on the overall mAP: 37.07\% $\rightarrow$ 36.97\% for 4-bit and 39.67\% $\rightarrow$ 39.70\% for 6-bit precision, respectively. 

The proposed regularization further increases critical performance on COCO panoptic dataset (\cref{tab:QAT_coco}~(bottom)) by 0.11\% and 0.34\% mAP for, correspondingly, 4-bit and 5-bit precision settings when compared to our PTQ results in~\cref{tab:PTQ_pano_fine}. The uniform PTQ quantization significantly underperforms in this setting.

Ablation study in~\cref{sec:reg_ablation} analyzes the impact of regularization strength. In addition, we show that our Fisher-trace regularization scheme that minimizes sharpness of the loss landscape improves model's \textit{test-time generalization}. Particularly, it outperforms a common heuristic approach when the critical objective $\mathcal{L}_F$ is simply added to the overall objective $\mathcal{L}_A$ during quantization-aware training.


\subsection{Qualitative Results of Fisher-trace Regularization}
\label{sec:QAT_qual}

\begin{table}[t]
\centering
\caption{Fisher trace of the critical objective when applied to DETR-R50 on COCO detection dataset. In this setting, the person category is considered as critical.}
\label{tab:fisher_trace}
\small
\begin{tabular}{cccc}
\toprule
Precision & Quant. scheme & Regularization & Fisher trace\\ 
\midrule
\multirow{3}{*}{4-bit}
& Uniform & No & 37.3K \\
&\ourcell Fisher-critical &\ourcell No &\ourcell 30.4K  \\
&\ourcell Fisher-critical &\ourcell Yes &\ourcell \textbf{14.9K} \\
\midrule
\multirow{3}{*}{6-bit}
 & Uniform & No & 88.9K\\
 &\ourcell Fisher-critical  &\ourcell No &\ourcell 18.2K\\
 &\ourcell Fisher-critical &\ourcell Yes &\ourcell \textbf{15.5K}\\
\bottomrule
\end{tabular}
\end{table}

To further show the effectiveness of the Fisher-trace regularization, we compute the Fisher trace of the critical objective on the quantized DETR model after QAT for the person category. We compare the Fisher trace of models with different quantization and training schemes in~\cref{tab:fisher_trace} that is estimated using 10,000 data points randomly sampled from the COCO detection dataset. 

Both the 4-bit and 6-bit uniform quantization settings lead to the largest Fisher trace on the critical objective, while our Fisher-aware quantization scheme helps to reduce the trace after QAT. Furthermore, the proposed regularization results in the lowest value. This observation confirms our analytical result in~\cref{ssec:cause}, where large Fisher trace indicates the least sharp local minima and, therefore, leads to inferior test-time generalization for critical categories.

\section{Conclusions}
\label{sec:con}

This work investigated the impact of quantization on the fine-grained performance of DETR-based object detectors. Motivated by safety concerns in practical applications, we formulated the critical-category objectives via the logit-label transformation of the corresponding categories. We empirically found that both the conventional PTQ and QAT cause disparate quantization effects.

We theoretically linked the disparate quantization effects with the sensitivity to the quantization weight perturbation and the sharpness of the loss landscape in the QAT. We characterized both derivations using the trace of the Fisher information matrix w.r.t. model weights. We proposed the Fisher-aware mixed-precision quantization scheme and Fisher-trace regularization to improve the critical-category performance of interest. We hope this work motivates future explorations on the fine-grained impacts of other compression methods in the computer vision area and a general machine learning research.

\section*{Acknowledgement}
We thank Panasonic and Berkeley Deep Drive for supporting this research.

\section*{Impact Statement}
This paper presents work whose goal is to advance the field of Machine Learning, specifically improving the finegrained performance of the critical categories of interest of the DETR model. In practical applications, the model's capability of correctly detecting each object category is not equally valuable. Some critical category appears to be more impactful to the general utility of the model, or leads to a more significant impact on the safety and trustworthiness of the application. We hope this work motivates future explorations on the fine-grained impacts of other compression methods in the computer vision area and a general machine learning research. 

\bibliography{main}

\begin{thebibliography}{44}
\providecommand{\natexlab}[1]{#1}
\providecommand{\url}[1]{\texttt{#1}}
\expandafter\ifx\csname urlstyle\endcsname\relax
  \providecommand{\doi}[1]{doi: #1}\else
  \providecommand{\doi}{doi: \begingroup \urlstyle{rm}\Url}\fi

\bibitem[Barocas et~al.(2019)Barocas, Hardt, and Narayanan]{barocas-hardt-narayanan}
Barocas, S., Hardt, M., and Narayanan, A.
\newblock \emph{Fairness and Machine Learning: Limitations and Opportunities}.
\newblock fairmlbook.org, 2019.
\newblock \url{http://www.fairmlbook.org}.

\bibitem[Bengio et~al.(2013)Bengio, L{\'e}onard, and Courville]{bengio2013estimatingSTE}
Bengio, Y., L{\'e}onard, N., and Courville, A.
\newblock Estimating or propagating gradients through stochastic neurons for conditional computation.
\newblock \emph{arXiv:1308.3432}, 2013.

\bibitem[Cai et~al.(2020)Cai, Gan, Wang, Zhang, and Han]{cai2019once}
Cai, H., Gan, C., Wang, T., Zhang, Z., and Han, S.
\newblock {Once-for-All}: Train one network and specialize it for efficient deployment.
\newblock In \emph{International Conference on Learning Representations (ICLR)}, 2020.

\bibitem[Carion et~al.(2020)Carion, Massa, Synnaeve, Usunier, Kirillov, and Zagoruyko]{carion2020end}
Carion, N., Massa, F., Synnaeve, G., Usunier, N., Kirillov, A., and Zagoruyko, S.
\newblock End-to-end object detection with transformers.
\newblock In \emph{Proceedings of the European Conference on Computer Vision (ECCV)}, 2020.

\bibitem[Choi et~al.(2018)Choi, Wang, Venkataramani, Chuang, Srinivasan, and Gopalakrishnan]{choi2018pact}
Choi, J., Wang, Z., Venkataramani, S., Chuang, P. I.-J., Srinivasan, V., and Gopalakrishnan, K.
\newblock {PACT}: Parameterized clipping activation for quantized neural networks.
\newblock \emph{arXiv:1805.06085}, 2018.

\bibitem[Cordts et~al.(2016)Cordts, Omran, Ramos, Rehfeld, Enzweiler, Benenson, Franke, Roth, and Schiele]{cityscapes}
Cordts, M., Omran, M., Ramos, S., Rehfeld, T., Enzweiler, M., Benenson, R., Franke, U., Roth, S., and Schiele, B.
\newblock The {C}ityscapes dataset for semantic urban scene understanding.
\newblock In \emph{Proceedings of the IEEE/CVF Conference on Computer Vision and Pattern Recognition (CVPR)}, 2016.

\bibitem[Ding et~al.(2019)Ding, Ding, Guo, and Han]{ding2019centripetal}
Ding, X., Ding, G., Guo, Y., and Han, J.
\newblock Centripetal {SGD} for pruning very deep convolutional networks with complicated structure.
\newblock In \emph{Proceedings of the IEEE/CVF Conference on Computer Vision and Pattern Recognition (CVPR)}, 2019.

\bibitem[Dong et~al.(2019)Dong, Yao, Gholami, Mahoney, and Keutzer]{dong2019hawq}
Dong, Z., Yao, Z., Gholami, A., Mahoney, M.~W., and Keutzer, K.
\newblock {HAWQ}: Hessian aware quantization of neural networks with mixed-precision.
\newblock In \emph{Proceedings of the IEEE International Conference on Computer Vision (ICCV)}, 2019.

\bibitem[Dong et~al.(2020)Dong, Yao, Arfeen, Gholami, Mahoney, and Keutzer]{dong2020hawq}
Dong, Z., Yao, Z., Arfeen, D., Gholami, A., Mahoney, M.~W., and Keutzer, K.
\newblock {HAWQ-V2}: Hessian aware trace-weighted quantization of neural networks.
\newblock \emph{Advances in neural information processing systems}, 2020.

\bibitem[Foret et~al.(2021)Foret, Kleiner, Mobahi, and Neyshabur]{foret2020sharpness}
Foret, P., Kleiner, A., Mobahi, H., and Neyshabur, B.
\newblock Sharpness-aware minimization for efficiently improving generalization.
\newblock In \emph{International Conference on Learning Representations (ICLR)}, 2021.

\bibitem[Geiger et~al.(2012)Geiger, Lenz, and Urtasun]{Geiger2012CVPR}
Geiger, A., Lenz, P., and Urtasun, R.
\newblock Are we ready for autonomous driving? {T}he {KITTI} vision benchmark suite.
\newblock In \emph{Proceedings of the IEEE/CVF Conference on Computer Vision and Pattern Recognition (CVPR)}, 2012.

\bibitem[Girshick(2015)]{girshickICCV15fastrcnn}
Girshick, R.
\newblock Fast {R-CNN}.
\newblock In \emph{Proceedings of the IEEE/CVF International Conference on Computer Vision (ICCV)}, 2015.

\bibitem[Good et~al.(2022)Good, Lin, Yu, Sieg, Fergurson, Zhe, Wieczorek, and Serra]{good2022recall}
Good, A., Lin, J., Yu, X., Sieg, H., Fergurson, M., Zhe, S., Wieczorek, J., and Serra, T.
\newblock Recall distortion in neural network pruning and the undecayed pruning algorithm.
\newblock In \emph{Advances in Neural Information Processing Systems}, 2022.

\bibitem[Gudovskiy et~al.(2021)Gudovskiy, Rigazio, Ishizaka, Kozuka, and Tsukizawa]{Gudovskiy_2021_CVPR}
Gudovskiy, D., Rigazio, L., Ishizaka, S., Kozuka, K., and Tsukizawa, S.
\newblock Auto{DO}: Robust autoaugment for biased data with label noise via scalable probabilistic implicit differentiation.
\newblock In \emph{Proceedings of the IEEE/CVF Conference on Computer Vision and Pattern Recognition (CVPR)}, 2021.

\bibitem[Guo et~al.(2022)Guo, Qiu, Leng, Gao, Zhang, Liu, Yang, Zhu, and Guo]{guo2022squant}
Guo, C., Qiu, Y., Leng, J., Gao, X., Zhang, C., Liu, Y., Yang, F., Zhu, Y., and Guo, M.
\newblock {SQ}uant: On-the-fly data-free quantization via diagonal {H}essian approximation.
\newblock In \emph{International Conference on Learning Representations (ICLR)}, 2022.

\bibitem[Han et~al.(2015)Han, Mao, and Dally]{han2015deep}
Han, S., Mao, H., and Dally, W.~J.
\newblock Deep compression: Compressing deep neural networks with pruning, trained quantization and huffman coding.
\newblock \emph{arXiv:1510.00149}, 2015.

\bibitem[Hinton et~al.(2015)Hinton, Vinyals, Dean, et~al.]{hinton2015distilling}
Hinton, G., Vinyals, O., Dean, J., et~al.
\newblock Distilling the knowledge in a neural network.
\newblock \emph{arXiv:1503.02531}, 2015.

\bibitem[Horowitz(2014)]{horowitz20141}
Horowitz, M.
\newblock 1.1 computing's energy problem (and what we can do about it).
\newblock In \emph{Proceedings of the IEEE International Solid-State Circuits Conference Digest of Technical Papers (ISSCC)}, 2014.

\bibitem[Kwon et~al.(2022)Kwon, Kim, Mahoney, Hassoun, Keutzer, and Gholami]{kwon2022fast}
Kwon, W., Kim, S., Mahoney, M.~W., Hassoun, J., Keutzer, K., and Gholami, A.
\newblock A fast post-training pruning framework for transformers.
\newblock In \emph{Advances in Neural Information Processing Systems}, 2022.

\bibitem[LeCun et~al.(1989)LeCun, Denker, and Solla]{lecun1989optimal}
LeCun, Y., Denker, J., and Solla, S.
\newblock Optimal brain damage.
\newblock \emph{Advances in Neural Information Processing Systems}, 1989.

\bibitem[Li et~al.(2022)Li, Zhang, Liu, Guo, Ni, and Zhang]{li2022dn}
Li, F., Zhang, H., Liu, S., Guo, J., Ni, L.~M., and Zhang, L.
\newblock {DN-DETR}: Accelerate {DETR} training by introducing query denoising.
\newblock In \emph{Proceedings of the IEEE/CVF Conference on Computer Vision and Pattern Recognition (CVPR)}, 2022.

\bibitem[Lin et~al.(2014)Lin, Maire, Belongie, Hays, Perona, Ramanan, Dollar, and Zitnick]{lin2014microsoft}
Lin, T.-Y., Maire, M., Belongie, S., Hays, J., Perona, P., Ramanan, D., Dollar, P., and Zitnick, L.
\newblock Microsoft {COCO}: Common objects in context.
\newblock In \emph{Proceedings of the European Conference on Computer Vision (ECCV)}, 2014.

\bibitem[Liu et~al.(2021)Liu, Cai, and Zhuang]{liu2021sharpness}
Liu, J., Cai, J., and Zhuang, B.
\newblock Sharpness-aware quantization for deep neural networks.
\newblock \emph{arXiv preprint arXiv:2111.12273}, 2021.

\bibitem[Liu et~al.(2022)Liu, Li, Zhang, Yang, Qi, Su, Zhu, and Zhang]{liu2022dab}
Liu, S., Li, F., Zhang, H., Yang, X., Qi, X., Su, H., Zhu, J., and Zhang, L.
\newblock {DAB}-{DETR}: Dynamic anchor boxes are better queries for {DETR}.
\newblock In \emph{International Conference on Learning Representations (ICLR)}, 2022.

\bibitem[Ly et~al.(2017)Ly, Marsman, Verhagen, Grasman, and Wagenmakers]{LY201740}
Ly, A., Marsman, M., Verhagen, J., Grasman, R.~P., and Wagenmakers, E.-J.
\newblock A tutorial on {F}isher information.
\newblock \emph{Journal of Mathematical Psychology}, 80:\penalty0 40--55, 2017.

\bibitem[Perronnin \& Dance(2007)Perronnin and Dance]{old-fisher}
Perronnin, F. and Dance, C.
\newblock Fisher kernels on visual vocabularies for image categorization.
\newblock In \emph{Proceedings of the IEEE Conference on Computer Vision and Pattern Recognition (CVPR)}, 2007.

\bibitem[Polino et~al.(2018)Polino, Pascanu, and Alistarh]{polino2018model}
Polino, A., Pascanu, R., and Alistarh, D.
\newblock Model compression via distillation and quantization.
\newblock In \emph{International Conference on Learning Representations (ICLR)}, 2018.

\bibitem[Redmon et~al.(2016)Redmon, Divvala, Girshick, and Farhadi]{redmon2016you}
Redmon, J., Divvala, S., Girshick, R., and Farhadi, A.
\newblock You only look once: Unified, real-time object detection.
\newblock In \emph{Proceedings of the IEEE/CVF Conference on Computer Vision and Pattern Recognition (CVPR)}, 2016.

\bibitem[Soen \& Sun(2024)Soen and Sun]{fim}
Soen, A. and Sun, K.
\newblock Tradeoffs of diagonal fisher information matrix estimators.
\newblock \emph{arXiv:2402.05379}, 2024.

\bibitem[Tian et~al.(2019)Tian, Shen, Chen, and He]{tian2019fcos}
Tian, Z., Shen, C., Chen, H., and He, T.
\newblock {FCOS}: Fully convolutional one-stage object detection.
\newblock In \emph{Proceedings of the IEEE/CVF International Conference on Computer Vision (ICCV)}, 2019.

\bibitem[Tran et~al.(2022)Tran, Fioretto, Kim, and Naidu]{tran2022pruning}
Tran, C., Fioretto, F., Kim, J.-E., and Naidu, R.
\newblock Pruning has a disparate impact on model accuracy.
\newblock In \emph{Advances in Neural Information Processing Systems}, 2022.

\bibitem[Wang et~al.(2022)Wang, Zhang, Cao, Shen, and Tao]{wang2022towards}
Wang, W., Zhang, J., Cao, Y., Shen, Y., and Tao, D.
\newblock Towards data-efficient detection transformers.
\newblock In \emph{Proceedings of the European Conference on Computer Vision (ECCV)}, 2022.

\bibitem[Wen et~al.(2016)Wen, Wu, Wang, Chen, and Li]{wen2016learning}
Wen, W., Wu, C., Wang, Y., Chen, Y., and Li, H.
\newblock Learning structured sparsity in deep neural networks.
\newblock In \emph{Advances in neural information processing systems}, 2016.

\bibitem[Wen et~al.(2017)Wen, Xu, Wu, Wang, Chen, and Li]{wen2017coordinating}
Wen, W., Xu, C., Wu, C., Wang, Y., Chen, Y., and Li, H.
\newblock Coordinating filters for faster deep neural networks.
\newblock In \emph{Proceedings of the IEEE International Conference on Computer Vision (ICCV)}, 2017.

\bibitem[Wu et~al.(2019)Wu, Dai, Zhang, Wang, Sun, Wu, Tian, Vajda, Jia, and Keutzer]{wu2019fbnet}
Wu, B., Dai, X., Zhang, P., Wang, Y., Sun, F., Wu, Y., Tian, Y., Vajda, P., Jia, Y., and Keutzer, K.
\newblock {FBNet}: Hardware-aware efficient convnet design via differentiable neural architecture search.
\newblock In \emph{Proceedings of the IEEE/CVF Conference on Computer Vision and Pattern Recognition (CVPR)}, 2019.

\bibitem[Xiao et~al.(2023)Xiao, Yang, Dong, Keutzer, Du, and Zhang]{xiao2022csq}
Xiao, L., Yang, H., Dong, Z., Keutzer, K., Du, L., and Zhang, S.
\newblock {CSQ}: Growing mixed-precision quantization scheme with bi-level continuous sparsification.
\newblock In \emph{Proceedings of the ACM/IEEE Design Automation Conference (DAC)}, 2023.

\bibitem[Yang et~al.(2020{\natexlab{a}})Yang, Tang, Wen, Yan, Hu, Li, Li, and Chen]{yang2020learning}
Yang, H., Tang, M., Wen, W., Yan, F., Hu, D., Li, A., Li, H., and Chen, Y.
\newblock Learning low-rank deep neural networks via singular vector orthogonality regularization and singular value sparsification.
\newblock In \emph{Proceedings of the IEEE/CVF Conference on Computer Vision and Pattern Recognition (CVPR) Workshops}, pp.\  678--679, 2020{\natexlab{a}}.

\bibitem[Yang et~al.(2020{\natexlab{b}})Yang, Wen, and Li]{Yang2020DeepHoyer}
Yang, H., Wen, W., and Li, H.
\newblock {DeepHoyer}: Learning sparser neural network with differentiable scale-invariant sparsity measures.
\newblock In \emph{International Conference on Learning Representations (ICLR)}, 2020{\natexlab{b}}.

\bibitem[Yang et~al.(2021)Yang, Duan, Chen, and Li]{yang2021bsq}
Yang, H., Duan, L., Chen, Y., and Li, H.
\newblock {BSQ}: Exploring bit-level sparsity for mixed-precision neural network quantization.
\newblock In \emph{International Conference on Learning Representations (ICLR)}, 2021.

\bibitem[Yang et~al.(2022)Yang, Yang, Gong, and Chen]{yang2021hero}
Yang, H., Yang, X., Gong, N.~Z., and Chen, Y.
\newblock {HERO}: Hessian-enhanced robust optimization for unifying and improving generalization and quantization performance.
\newblock In \emph{Proceedings of the ACM/IEEE Design Automation Conference (DAC)}, 2022.

\bibitem[Yang et~al.(2023)Yang, Yin, Shen, Molchanov, Li, and Kautz]{yang2021nvit}
Yang, H., Yin, H., Shen, M., Molchanov, P., Li, H., and Kautz, J.
\newblock Global vision transformer pruning with {H}essian-aware saliency.
\newblock In \emph{Proceedings of the IEEE/CVF Conference on Computer Vision and Pattern Recognition (CVPR)}, 2023.

\bibitem[Yao et~al.(2021)Yao, Dong, Zheng, Gholami, Yu, Tan, Wang, Huang, Wang, Mahoney, et~al.]{yao2021hawq}
Yao, Z., Dong, Z., Zheng, Z., Gholami, A., Yu, J., Tan, E., Wang, L., Huang, Q., Wang, Y., Mahoney, M., et~al.
\newblock {HAWQ-V3}: Dyadic neural network quantization.
\newblock In \emph{International Conference on Machine Learning (ICML)}, 2021.

\bibitem[Yuan et~al.(2022)Yuan, Xue, Chen, Wu, and Sun]{yuan2022ptq4vit}
Yuan, Z., Xue, C., Chen, Y., Wu, Q., and Sun, G.
\newblock {PTQ4ViT}: Post-training quantization for vision transformers with twin uniform quantization.
\newblock In \emph{Proceedings of the European Conference on Computer Vision (ECCV)}, pp.\  191--207, 2022.

\bibitem[Zhu et~al.(2021)Zhu, Su, Lu, Li, Wang, and Dai]{zhu2020deformable}
Zhu, X., Su, W., Lu, L., Li, B., Wang, X., and Dai, J.
\newblock Deformable {DETR}: Deformable transformers for end-to-end object detection.
\newblock In \emph{International Conference on Learning Representations (ICLR)}, 2021.

\end{thebibliography}
\bibliographystyle{want_icml2024}

\newpage
\appendix
\onecolumn

\appendix

We provide supplementary materials in the following appendices. Specifically,~\cref{ap:derive} provides detailed derivation of~\cref{equ:grad_norm}.~\cref{sec:PTQ_ablation} further analyzes how the overall and critical-category objectives from~\cref{equ:ILP} impact the overall performance metric, and contains additional CityScapes experimental results.~\cref{sec:reg_ablation} conducts an ablation study on the impact of regularization strength $\lambda$ for our Fisher-trace regularization and verifies the design choice of using our regularizer instead of a simple summation of overall and critical-category objectives for the QAT optimization.~\cref{sec:PTQ_detail} provides additional qualitative visualizations of the layer-wise Fisher-aware sensitivity and the corresponding quantization assignments derived using the proposed method.

\section{Detailed Derivation of~\cref{equ:grad_norm}}
\label{ap:derive}

The first-order Taylor expansion of the perturbed loss $\mathcal{L}_F(q(\vtheta)+\vepsilon)$ in~\cref{equ:sharp} is
\begin{equation}
    \label{equ:perturb_taylor}
    \mathcal{L}_F(q(\vtheta)+\vepsilon) \approx \mathcal{L}_F(q(\vtheta)) + \vepsilon^T \partial \mathcal{L}_F(q(\vtheta)) / \partial \vtheta.
\end{equation}
By substituting~\cref{equ:perturb_taylor} into~\cref{equ:sharp}, the maximization can be simplified as
\begin{equation}
\label{equ:sharp_taylor}
    \mathcal{S}(q(\vtheta)) = \max_{\|\vepsilon\|_2\leq\rho}\mathcal{L}_F(q(\vtheta) + \vepsilon)-\mathcal{L}_F(q(\vtheta)) \approx \max_{||\vepsilon||_2\leq\rho} \vepsilon^T \partial \mathcal{L}_F(q(\vtheta)) / \partial \vtheta.
\end{equation}
Note that both the $\vepsilon$ and $\partial \mathcal{L}_F(q(\vtheta)) / \partial \vtheta$ are vectors with the same dimensions as weight vector $\vtheta$. Then, their inner product achieves the maximum when they are parallel. Therefore, we can solve the maximization in~\cref{equ:sharp_taylor} as
\begin{equation}
\begin{split}
\label{equ:derive}
    \mathcal{S}(q(\vtheta)) &\approx \max_{||\vepsilon||_2\leq\rho} \vepsilon^T \partial \mathcal{L}_F(q(\vtheta)) / \partial \vtheta \\
    &= \frac{\rho}{\| \partial \mathcal{L}_F(q(\vtheta)) / \partial \vtheta \|_2} \frac{\partial \mathcal{L}_F(q(\vtheta))}{\partial \vtheta^T} \frac{\partial \mathcal{L}_F(q(\vtheta))}{\partial \vtheta} \\
    &\propto \frac{\partial \mathcal{L}_F(q(\vtheta))}{\partial \vtheta^T} \frac{\partial \mathcal{L}_F(q(\vtheta))}{\partial \vtheta} = \tr(\mathcal{I}),
\end{split}
\end{equation}
which is the final approximation of the loss landscape sharpness in~\cref{equ:grad_norm}.

\section{Ablation Study on Fisher-aware Quantization}
\label{sec:PTQ_ablation}

~\cref{tab:PTQ_COCO_fine} contains only the critical-category metrics. Here we report the overall mAPs in \cref{tab:4PTQ_det_over,tab:6PTQ_det_over} to show the impact on the overall performance. 
In general, Fisher-critical quantization scheme leads to comparable overall metrics with the Fisher-overall scheme, and they both are significantly higher than the conventional uniform and HAWQ-V2 quantization schemes. In some cases, the improvement of critical-category metrics with the proposed scheme also improves the overall performance. 
This indicates that the addition of such critical-category objective in the sensitivity analysis can be useful for increasing the overall performance as well. This is an interesting direction for future work.

\begin{table}[h]
\centering
\small
\caption{Overall mAP on COCO detection dataset for 4-bit precision budget, \%.}
\label{tab:4PTQ_det_over}
\begingroup
    \setlength{\tabcolsep}{2pt}
\begin{tabular}{c|c|cccc}
\toprule
Model & Uniform & Fisher-overall & \ourcell Fisher-person & \ourcell Fisher-animal & \ourcell Fisher-indoor \\
\midrule
DETR-R50 & 36.7 & \textbf{37.12}\tiny$\pm$0.1 &\ourcell 37.1\tiny$\pm$0.1 &\ourcell 37.0\tiny$\pm$0.1 &\ourcell 36.99\tiny$\pm$0.1        \\
DETR-R101 & 37.4 & \textbf{38.26} &\ourcell 38.22 &\ourcell 37.97 &\ourcell 38.24 \\
DAB DETR-R50 & 22.7 & 24.42 &\ourcell 25.28 &\ourcell \textbf{25.84} &\ourcell 24.08 \\
Deformable DETR-R50 & 28.8 & 44.1 &\ourcell \textbf{44.5} &\ourcell 44.1 &\ourcell \textbf{44.5} \\
\bottomrule   
\end{tabular}
\endgroup
\end{table}

\begin{table}[h]
\centering
\small
\caption{Overall mAP on COCO detection dataset for 6-bit precision budget, \%.}
\label{tab:6PTQ_det_over}
\begingroup
    \setlength{\tabcolsep}{2pt}
\begin{tabular}{c|c|cccc}
\toprule
Model & Uniform & Fisher-overall & \ourcell Fisher-person & \ourcell Fisher-animal & \ourcell Fisher-indoor \\
\midrule
DETR-R50 & 39.4 & 39.57\tiny$\pm$0.1 &\ourcell \textbf{39.67}\tiny$\pm$0.1 &\ourcell 39.60\tiny$\pm$0.0 &\ourcell 39.61\tiny$\pm$0.1        \\
DETR-R101 & 39.2 & 41.8 &\ourcell 41.8 &\ourcell \textbf{42.1} &\ourcell \textbf{42.1} \\
DAB DETR-R50 & 28.00 & 27.20 &\ourcell \textbf{28.42} &\ourcell 27.30 &\ourcell 27.94 \\
Deformable DETR-R50 & 47.8 & 48.1 &\ourcell \textbf{48.5} &\ourcell 48.1 &\ourcell \textbf{48.5} \\
\bottomrule   
\end{tabular}
\endgroup
\end{table}

Finally, we provide additional results of the Fisher-aware quantization scheme with the CityScapes dataset in~\cref{tab:cityscape}. We perform object detection task with DETR model on the CityScapes dataset following the settings of~\cite{wang2022towards}\footnote{\url{https://github.com/encounter1997/DE-DETRs}}. We can observe the same trend that the proposed Fisher-overall scheme significantly outperforms uniform quantization, whereas Fisher-critical scheme further improves the performance of the corresponding critical categories. 

\begin{table}[h]
\centering
\caption{Critical-category mAP on CityScapes for DETR-R50, \%.}
\label{tab:cityscape}
\begin{tabular}{ccccccc}
\toprule
 \multirow{2}{*}{Precision} & \multirow{2}{*}{\shortstack{Quant.\\scheme}} &  \multirow{2}{*}{Overall}  & \multicolumn{4}{c}{Critical category} \\
&  &  &  Construct & Object &  Human   &  Vehicle \\
\midrule
FP & -  & 11.7           & 8.7             & 17.8            & 18.0       & 19.0      \\
\midrule
\multirow{3}{*}{4-bit}  & Uniform &  5.2       & 3.6       & 8.7        & 8.8      & 9.2        \\
& Fisher-overall & 8.8      &  5.5      & 12.6       & 13.8       & 14.6          \\
& \ourcell Fisher-critical   &  \ourcell \textbf{9.0}          &\ourcell \textbf{6.5}             &\ourcell \textbf{13.7}            &\ourcell \textbf{14.0}                          &\ourcell \textbf{14.7}        \\
\bottomrule
\end{tabular}
\end{table}


\section{Ablation Study on Fisher-trace Regularization}
\label{sec:reg_ablation}

We start with the discussion about the impact of regularization strength $\lambda$ on the overall and the critical performance in the QAT. Similarly to previous work on the regularized training~\cite{yang2021hero}, $\lambda$ controls the tradeoff between the overall performance and the generalization gap for the critical objective.
~\cref{tab:reg_abl} shows the overall and critical mAP during training if we set $\lambda$ to a smaller value e.g., 1e-3. It can be seen that the Fisher trace regularization significantly improves critical mAP during epoch range from 20 to 30 (up to 0.5\%). As the training progresses towards convergence, the critical-category performance drops while the overall performance increases which indicating the occurrence of overfitting.

\begin{table}[t!]
\centering
\caption{QAT performance of DETR-R50 on COCO detection dataset. The person category is considered as critical and 4-bit Fisher-critical quantization scheme is applied. The mAP metrics at each epoch are reported using overall/critical format, \%.}
\label{tab:reg_abl}
\begin{tabular}{c|ccccc}
\toprule
$\lambda$ & Epoch 10 & Epoch 20 & Epoch 30 & Epoch 40 & Epoch 50 \\
\midrule
0 & 36.8/34.8 & 36.7/34.7 & 36.9/34.9 & 37.3/35.1 & 37.2/35.2 \\
1e-3 & 36.5/34.5 & 36.9/\textbf{35.2} & 36.8/\textbf{35.3} & 37.1/35.0 & 37.2/35.1 \\
\bottomrule   
\end{tabular}
\end{table}

However, setting the $\lambda$ too large (e.g., 5e-3) during the initial epochs of the QAT process significantly affects the convergence of the overall training objective. 
These observations indicate that during the QAT process, a smaller regularization is needed initially to facilitate convergence, while a larger regularization is needed towards the end to prevent the overfitting.
To address this in our work we utilize a linear scheduling of the regularization strength as discussed in ~\cref{sec:exp_setup}, which can be formulated as $\lambda = \max\left[\lambda_0, \lambda_T t/T \right]$, where $t$ is the current epoch, $T$ is the total number of epochs, and $\lambda_0, \lambda_T$ are the initial and final regularization strengths, respectively. This scheme leads to higher results in~\cref{tab:QAT_coco,tab:PTQ_pano_fine}.

Finally, we verify the necessity of applying Fisher-trace regularization during QAT. Specifically, we compare to a common heuristic approach when the critical objective $\mathcal{L}_F$ is simply added to the overall objective $\mathcal{L}_A$.
For the Fisher-trace regularization, the motivation comes from our Claim 2 in~\cref{ssec:cause}, where the QAT gap is caused by the \textit{sharp loss landscape}, which leads to a \textit{poor generalization}.~\cref{tab:sgd} results confirm that the test-time generalization cannot be improved by the addition of critical objectives to the training loss. 

\begin{table}[t!]
\centering
\caption{QAT performance of DETR-R50 model on COCO detection (left) and panoptic (right) datasets. All models are quantized using the Fisher-critical scheme with 4-bit budget for detection and 5-bit budget for panoptic dataset, respectively.}
\label{tab:sgd}
\begin{tabular}{ccc||ccc}
\toprule
\multirow{2}{*}{\shortstack{QAT\\objective}} & \multirow{2}{*}{Overall} & \multirow{2}{*}{Person} &
\multirow{2}{*}{\shortstack{QAT\\objective}} & \multirow{2}{*}{Overall} & \multirow{2}{*}{Person} \\
& & & & & \\
\midrule
overall & 37.07\tiny$\pm$0.07 & 35.56\tiny$\pm$0.08 & overall & 36.08\tiny$\pm$0.07 & 19.05\tiny$\pm$0.09 \\
overall+critical & 37.11\tiny$\pm$0.04 & 35.39\tiny$\pm$0.06 & overall+critical & 36.07\tiny$\pm$0.05 & 19.19\tiny$\pm$0.10 \\
\rowcolor[rgb]{1,0.808,0.576}
Fisher reg. & 36.97\tiny$\pm$0.06 & \textbf{35.75}\tiny$\pm$0.04 & Fisher reg. & \textbf{36.12}\tiny$\pm$0.06 & \textbf{19.39}\tiny$\pm$0.11 \\
\bottomrule
\end{tabular}
\end{table}

\section{Qualitative Results of Fisher-aware Quantization}
\label{sec:PTQ_detail}

\subsection{Fisher-aware Sensitivity vs. Quantization Assignments}
\label{sec:PTQ_sens}

\begin{figure*}[h!]
    \centering
    \includegraphics[width=1\textwidth]{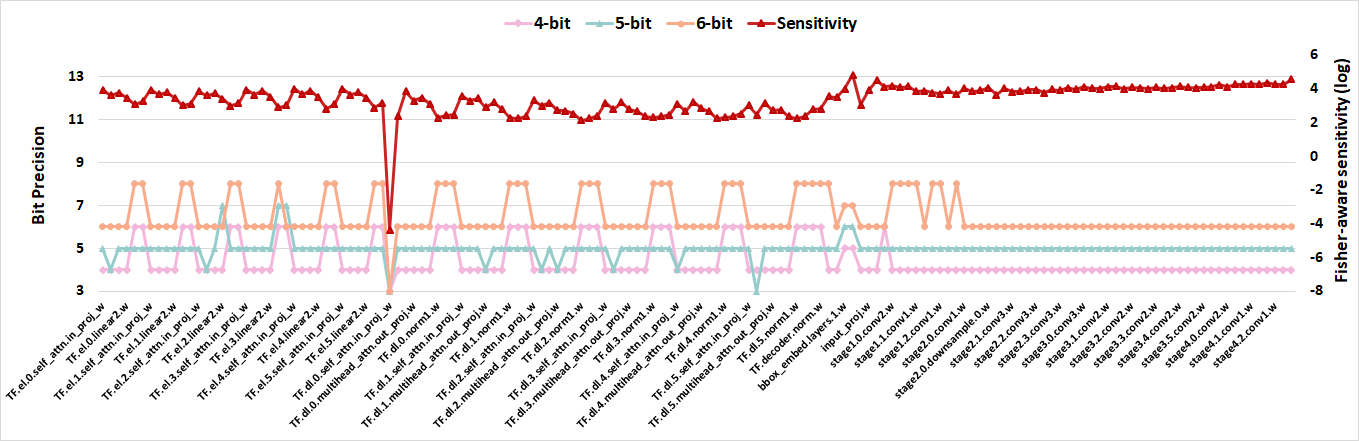}
    \caption{Bit precision vs. layer-wise sensitivity for DETR-R50 on COCO panoptic dataset. There is a clear correlation between the number of bits and our sensitivity metric.}
    \label{fig:Segm-DetrR50-sens}
\end{figure*}

\begin{figure*}[h!]
    \centering
    \includegraphics[width=1\textwidth]{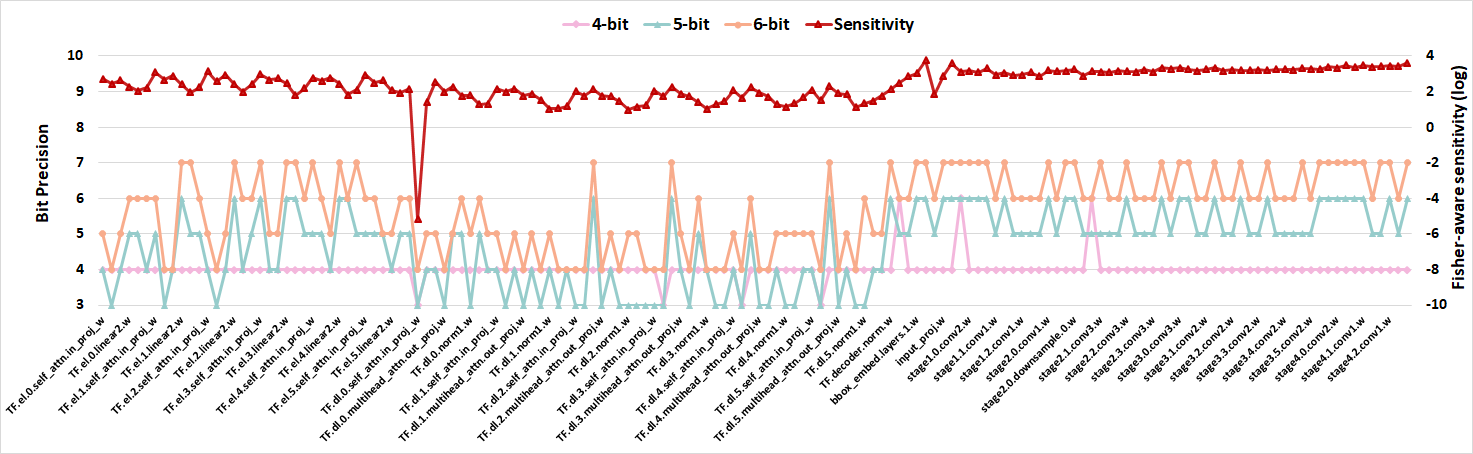}
    \includegraphics[width=1\textwidth]{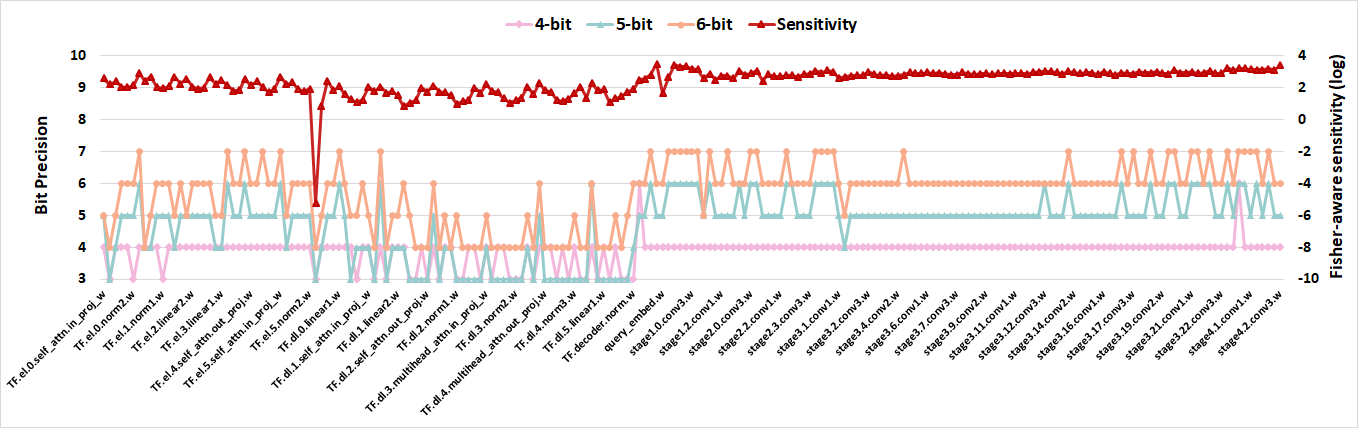}
    \includegraphics[width=1\textwidth]{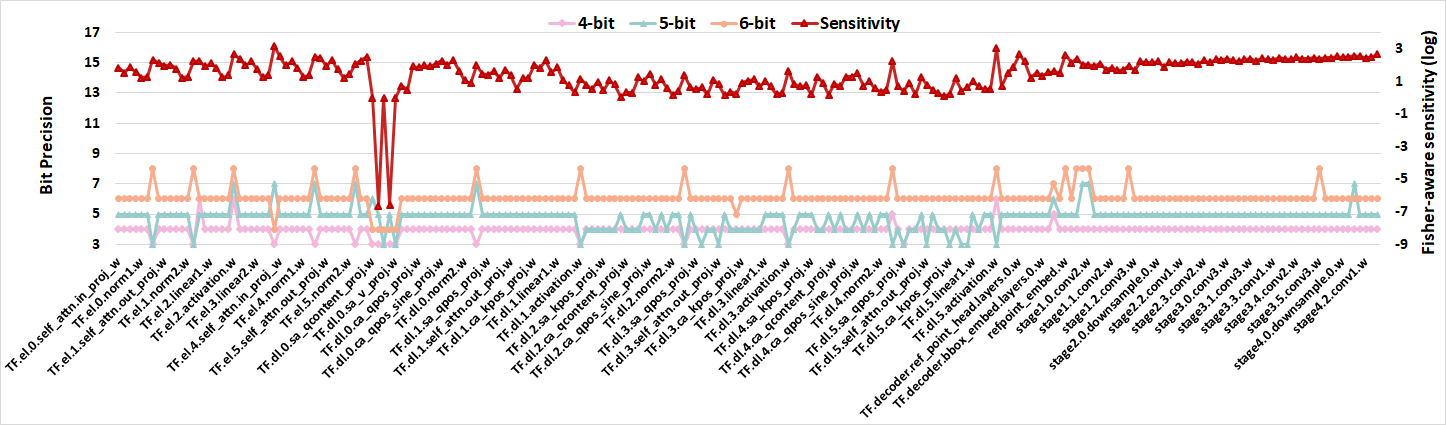}
    \includegraphics[width=1\textwidth]{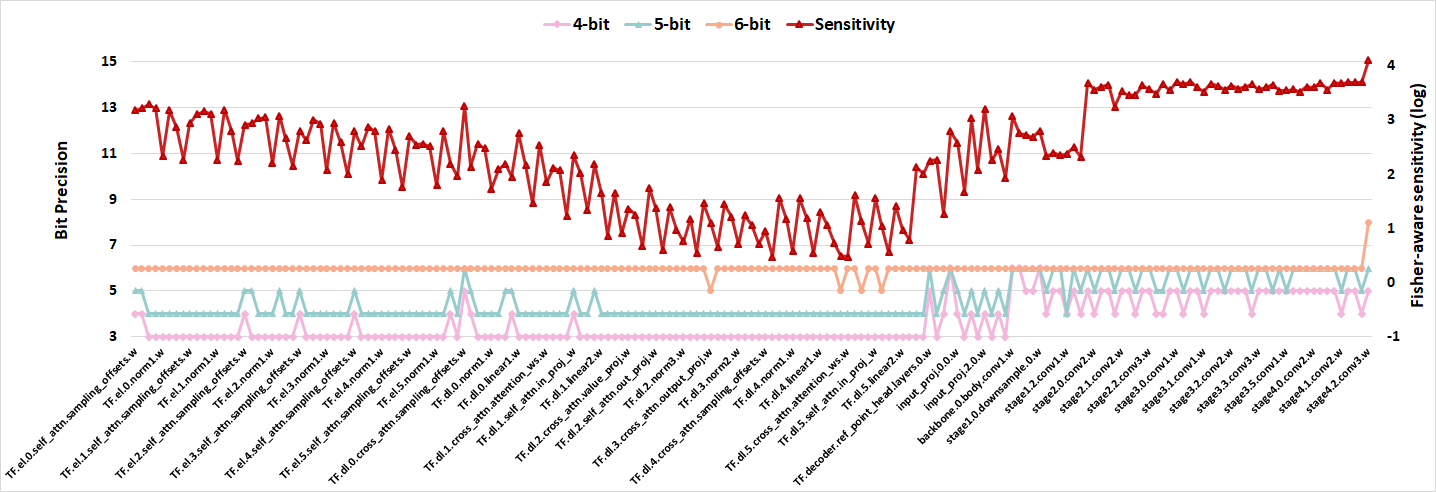}
    \caption{Bit precision vs. layer-wise sensitivity for DETR-R50, DETR-R101, DAB DETR-R50 and Deformable DETR-R50 on COCO detection dataset, respectively.}
    \label{fig:DetrR50-sen_pre}
\end{figure*}

\begin{figure*}[]
    \centering
    \includegraphics[width=1\textwidth]{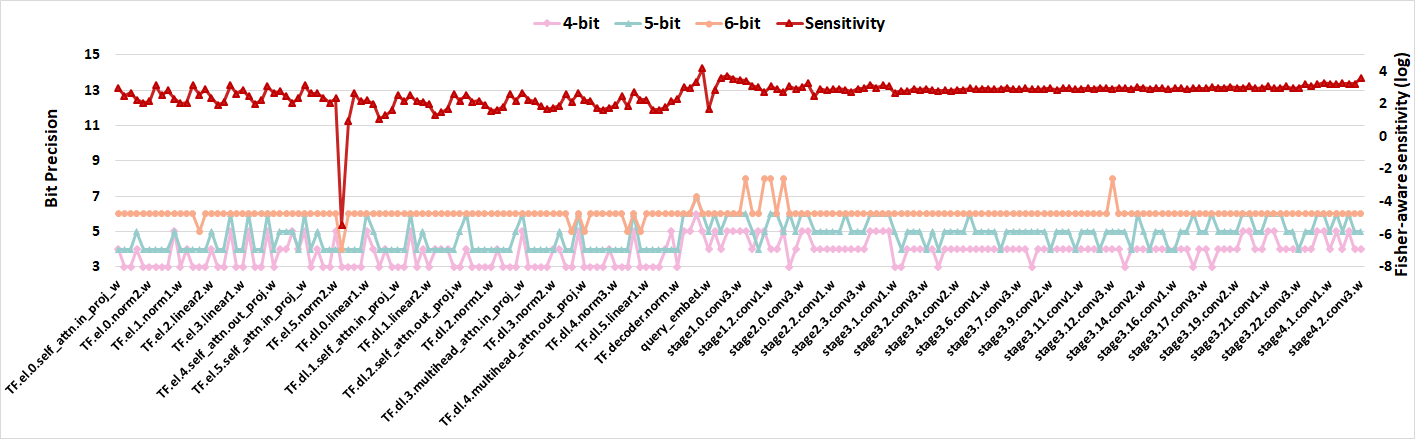}
    \caption{Bit precision vs. layer-wise sensitivity for DETR-R101 on COCO panoptic dataset.}
    \label{fig:Segm-DetrR50-sen_pre}
\end{figure*}

We illustrate the Fisher-aware sensitivity and the corresponding quantization assignments for a Fisher-overall scheme from~\cref{tab:PTQ_pano_fine} when applied to the DETR-R50 model on the COCO panoptic dataset in~\cref{fig:Segm-DetrR50-sens}. As shown in the visualization, the backbone layers demonstrate a relatively stable sensitivity distribution, while the transformer encoder and decoder layers show sensitivity distribution with high variance. This is expected given the different functionalities of transformer layers within an attention block~\cite{carion2020end}. Then, the quantization assignments 
are performed with the clear correlation between the sensitivity magnitude for each layer and the budget constraints when solving the ILP from~\cref{equ:ILP}. Additionally, we visualize DETR-R50, DETR-R101, DAB DETR-R50, and Deformable DETR-R50 models on the COCO detection dataset in \cref{fig:DetrR50-sen_pre}, and, additionally, draw DETR-R101 on the COCO panoptic dataset in~\cref{fig:Segm-DetrR50-sen_pre}.

\subsection{Overall vs. Critical-category Objectives for Quantization}
\label{sec:PTQ_diff}

\begin{figure*}[h!]
    \centering
    \includegraphics[width=1\textwidth]{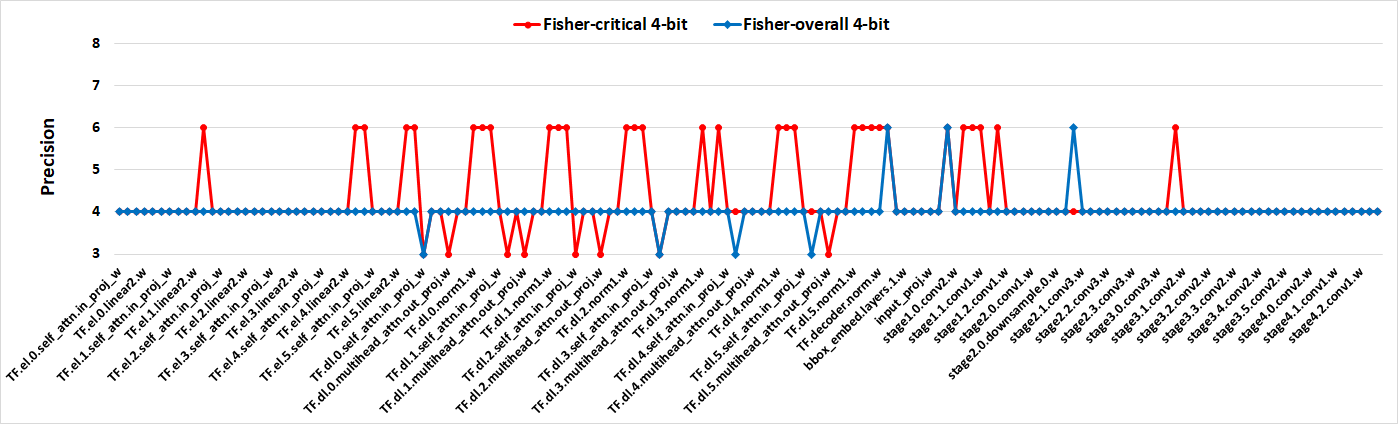}
    \caption{Comparison of Fisher-critical and Fisher-overall assignments for DETR-R50 on COCO detection dataset when applied to the person category. Our critical objective leads to a significant change in the precision assigned to detector's layers.}
    \label{fig:DetrR50-assign}
\end{figure*}

\begin{figure*}[]
    \centering
    \includegraphics[width=1\textwidth]{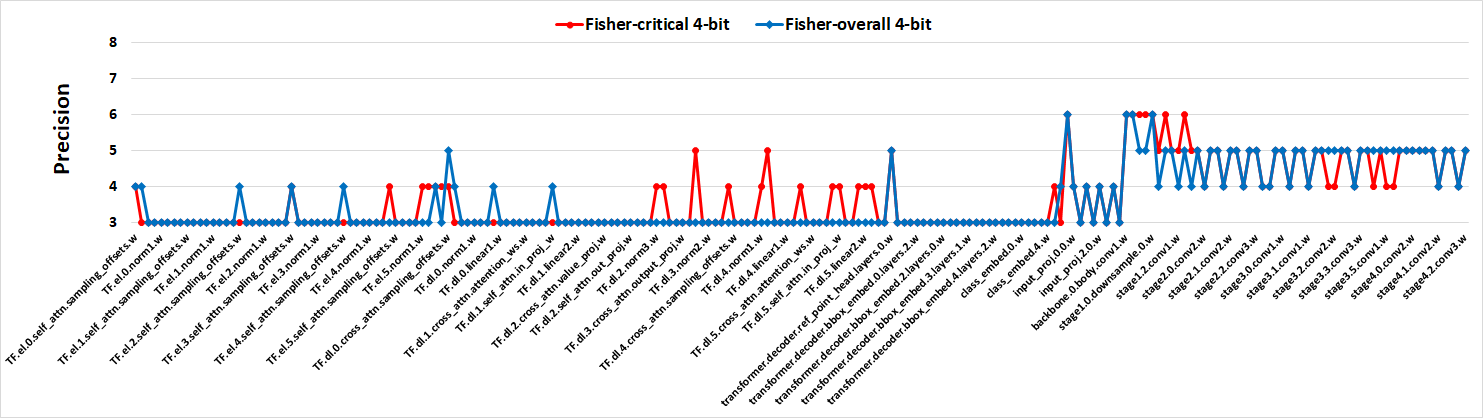}
    \caption{Comparison of Fisher-critical and Fisher-overall assignments for Deformable DETR-R50 on COCO detection dataset when applied to \textbf{person category}.}
    \label{fig:DetrR50-pre}
\end{figure*}

\begin{figure*}[]
    \centering
    \includegraphics[width=1\textwidth]{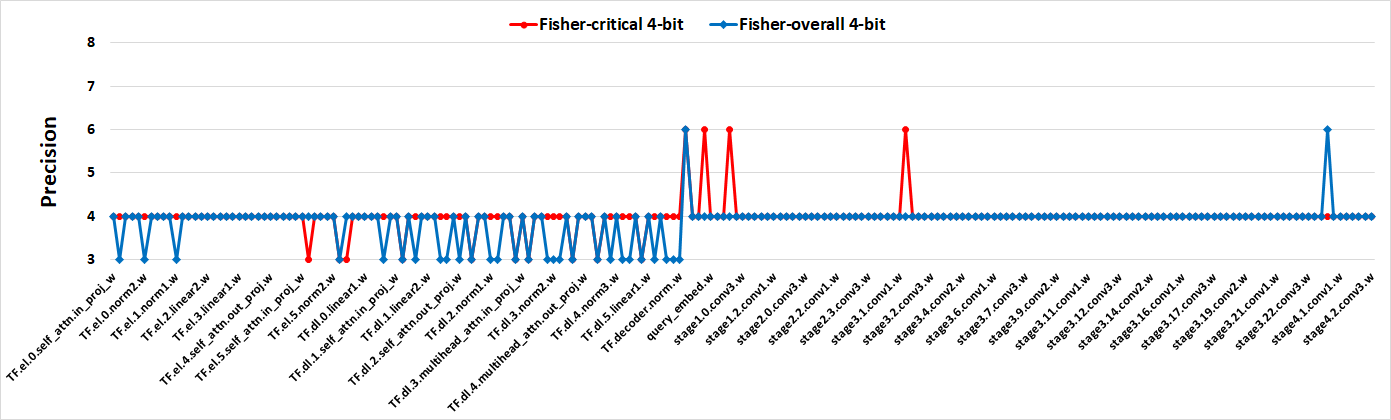}
    \includegraphics[width=1\textwidth]{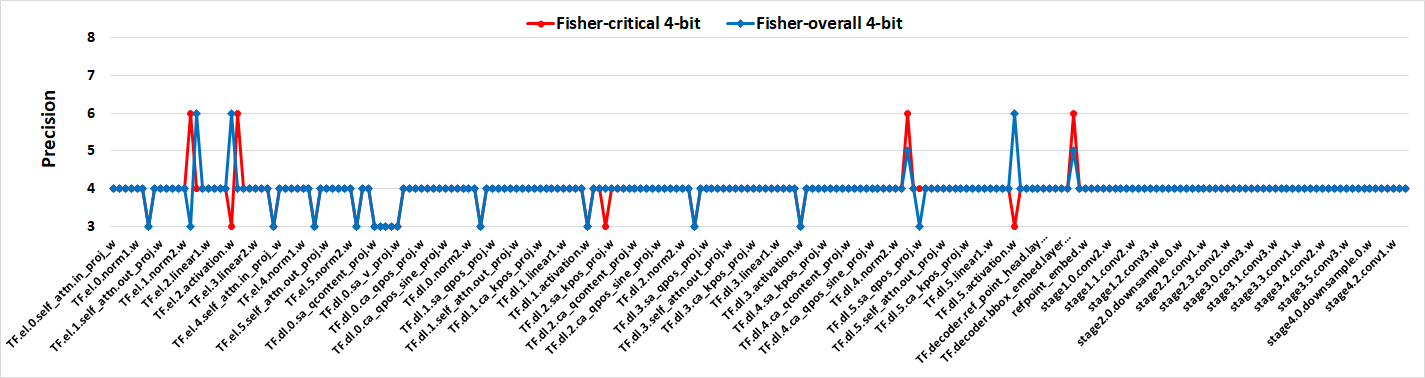}
    \caption{Comparison of Fisher-critical and Fisher-overall assignments for DETR-R101 and DAB DETR-R50 on COCO detection dataset when applied to \textbf{indoor} and \textbf{animal} categories, respectively.}
    \label{fig:DetrR101-pre}
\end{figure*}

\begin{figure*}[]
    \centering
    \includegraphics[width=1\textwidth]{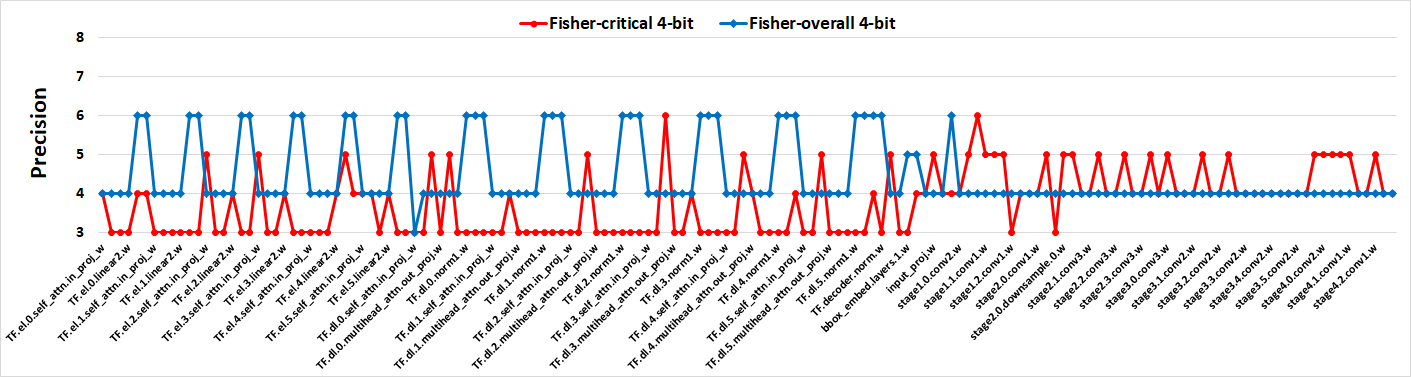}
    \includegraphics[width=1\textwidth]{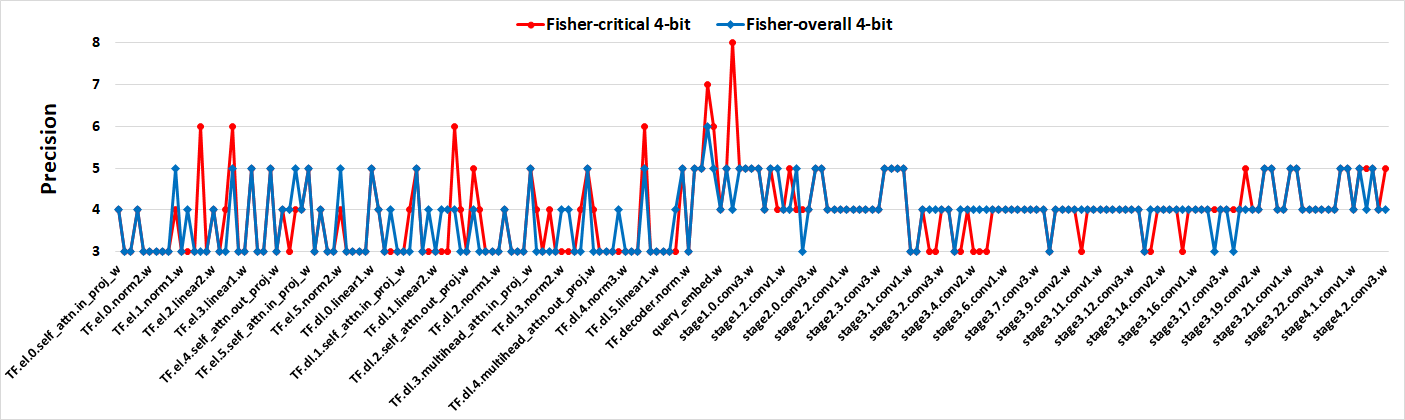}
    \caption{Comparison of Fisher-critical and Fisher-overall assignments for DETR-R50 and DETR-R101 on COCO panoptic dataset when applied to \textbf{animal} and \textbf{person} categories, respectively.}
    \label{fig:Segm-DetrR50-pre}
\end{figure*}

Next, we visually compare mixed-precision quantization assignments in our Fisher-aware scheme when the conventional "overall" objective or the proposed "critical-category" objective are employed.~\cref{fig:DetrR50-assign} compares the assignments for the DETR-R50 model on COCO detection dataset when applied to the person category that corresponds to the quantitative result in~\cref{tab:PTQ_COCO_fine} (top). As shown in the figure, the inclusion of critical objective into the ILP leads to a \textit{significant change in the precision assigned to certain layers}. In particular, our Fisher-critical scheme has more peaks and lows than a more smooth conventional scheme. This illustrates high sensitivity of layers to the critical-category objective. By adding the objectives of interest, it is possible to \textit{improve model's quantization at the fine-grained level}. Additionally,~\cref{fig:DetrR50-pre,fig:DetrR101-pre,fig:Segm-DetrR50-pre} compare the assignments for different models and critical categories reported in~\cref{tab:PTQ_COCO_fine,tab:PTQ_pano_fine}.

\end{document}